\documentclass{article}

\usepackage[preprint]{neurips_2021}

\usepackage{microtype}
\usepackage{graphicx}
\usepackage{subfigure}
\usepackage{booktabs} 
\usepackage{amssymb}
\usepackage{amsmath}
\usepackage{xcolor}
\usepackage{amsthm}
\usepackage{mathtools}

\usepackage[utf8]{inputenc} 
\usepackage[T1]{fontenc}    
\usepackage{hyperref}       
\usepackage{url}            
\usepackage{amsfonts}       
\usepackage{nicefrac}       

\newtheorem{theorem}{Theorem}
\newtheorem*{theorem*}{Theorem}

\title{Identification of Subgroups With Similar Benefits in Off-Policy Policy Evaluation}

\author{%
  Ramtin Keramati\\
  Institute of Computational and Mathematical Engineering\\
  Stanford University\\
  \texttt{keramati@cs.stanford.edu} \\
  \And 
  Omer Gottesman\\
  Department of Computer Science\\
  Brown University\\
  \texttt{ogottesm@cs.brown.edu}
  \AND
  Leo Anthony Celi\\
  Institute for Medical Engineering \& Science\\
  Massachusetts Institute of Technology\\
  \And
  Finale Doshi-Velez\\
  Department of Computer Science\\
  Harvard University\\
  \And
  Emma Brunskill\\
  Department of Computer Science\\
  Stanford University
}

\begin{document}

\maketitle

\begin{abstract}
  Off-policy policy evaluation methods for sequential decision making can be used to help identify if a proposed decision policy is better than a current baseline policy. However, a new decision policy may be better than a baseline policy for some individuals but not others. This has motivated a push towards personalization and accurate per-state estimates of heterogeneous treatment effects (HTEs). Given the limited data present in many important applications, individual predictions can come at a cost to accuracy and confidence in such predictions.  We develop a method to balance the need for personalization with confident predictions by identifying subgroups where it is possible to confidently estimate the expected difference in a new decision policy relative to a baseline. We propose a novel loss function that accounts for uncertainty during the subgroup partitioning phase. In experiments, we show that our method can be used to form accurate predictions of HTEs where other methods struggle. 
\end{abstract}

\section{Introduction}

Recent advances in technology and regulations around them
have enabled the collection of an unprecedented amount of data of past decisions
and outcomes in different domains such as health
care, recommendation systems, and education. This offers a
unique opportunity to learn better decision-making policies
using the observational data.
Off-policy policy evaluation (OPE) is concerned with estimating
the value of a proposed policy (\textit{evaluation policy}) using the data
collected under a different policy (\textit{behaviour policy}). 
Estimating the value of an evaluation
policy before deployment is essential,
especially when interacting with the environment is expensive, risky, or
unethical, such as in health care ~\citep{gottesman2019guidelines}. Fortunately, the reinforcement learning (RL) community has developed different
methods and theories focused on OPE, e.g. \citet{jiang2016doubly, thomas2016ope, kallus2020double}.

OPE has been used extensively in the
literature to demonstrate the superiority of a proposed
evaluation policy relative to the baseline (behaviour) policy e.g.~\citet{komorowski2018artificial};
however, the evaluation
policy may be better than the behaviour policy for some individuals
but not others. Hence, only looking at the estimated value of the
evaluation policy before deployment, may be misleading. In the non-sequential
setting, a growing literature has focused on personalization and estimation of heterogeneous treatment effect (HTE), the
individual-level differences in potential outcomes under the
proposed evaluation policy versus the behaviour policy~\citep{athey2019generalized, nie2017quasi}. 

However, due to the limited availability of data and a long horizon, the goal of personalization for each individual can be unrealistic, especially in
sequential settings. In practice, individual predictions of treatment effects in these settings can be inaccurate and highly uncertain, providing no actionable information. 

In this paper, we aim to provide actionable information
to domain experts. Specifically, we ask "\textit{What subgroups of individuals can we confidently predict that will be significantly benefited or harmed by adopting the evaluation policy?}". Asking this question instead of "\textit{What is the treatment effect for each individual?}"
allows us to group individuals that have
similar treatment effects together, pool the data,
and make predictions that are both more accurate and confident. This
information can be used by a domain expert (human-in-the-loop) to make informed
decisions before deploying the RL system, or be used for the interpretability
of the OPE. For example, a clinician can take a look at the groups and decide if
the predicted benefit or harm is in accordance with their clinical intuition.

We propose using a loss function similar to non-sequential setting \citep{athey2016recursive} and greedily minimize it with recursive partitioning, similar to the classification and regression trees (CART).
Estimation of the naive loss function can be too noisy and often results
in over-splitting, yielding too many
subgroups and inaccurate
or uncertain prediction. We make two main changes to the naive loss function. First we mitigate the issue of noisy estimation by proposing a novel upper bound to the loss function that is stable and can be efficiently calculated.


Additionally, by taking into account what clinicians believe is relevant for decision making, we incorporate a regularization term that incentivizes recursive partitioning to find subgroups with relevant treatment effects. This term can be directly specified by a clinician, and allows for better incorporation of medical experts into the evaluation process.
For example, a clinician may consider an increase of 10\% in survival rate relevant, so only subgroups with a confident prediction of 10\% decrease or increases in survival rate will provide actionable information. 
Combining these two additions, we propose a new loss function that can be efficiently computed.

On a simulated example of sepsis management \citep{oberst2019counterfactual} we show how our proposed method can be used to find subgroups with significant treatment effect, providing
more accurate and confident predictions.  
Additionally, we apply our method to the sepsis cohort of the MIMIC III ICU dataset \citep{johnson2016mimic}, and illustrate how it can be used to identify subgroups in which a new decision policy may 
be beneficial or harmful relative to the standard approach.
We also investigate the interpretability of our findings through a discussion with an ICU intensivist.

\vspace{-5pt}
\section{Related Work}
\vspace{-5pt}
Estimating the value of a new decision policy
arise in many different applications, such
as personalized medicine \citep{obermeyer2016predicting}, bandits
\citep{dudik2011doubly} and sequential decision
makings \citep{thomas2016ope}.
The RL community has developed different methods and theories
for off-policy policy evaluation (OPE) in sequential setting. These methods mostly
fall into different categories: importance sampling \citep{precup2000eligibility}, model based and doubly robust methods \citep{dudik2011doubly, thomas2016ope, jiang2016doubly}. These methods can be used along with
our algorithm to estimate group treatment effect for a
particular group; however,
these methods do not offer a way to perform partitioning. 

In non-sequential settings, a growing number of literature
seek to estimate heterogeneous treatment effect (HTE) using
different approaches.~\citet{imai2014covariate} 
uses the LASSO to estimates the effect of treatments,
\citet{shalit2017estimating}
uses neural networks and offers generalization bound
for individual treatment effect (ITE). \citet{nie2017quasi} proposed two step estimation procedure
using double machine learning and orthogonal moments \citep{chernozhukov2018double} that can be applied on
observational data to infer HTEs, and recently, \citet{lee2020robust}
suggests a robust partitioning algorithm by inducing homogeneity in 
groups.
However, these 
methods were developed for non-sequential
settings and na\"ively applying them to sequential setting
will result in predictions with low accuracy and high
uncertainty. Our work draws close
parallel to methods using recursive partitioning to estimate
HTEs~\citep{athey2016recursive, athey2019generalized},
but those works suffer from over-splitting the feature space in
sequential setting due to noisy estimation of the loss
function. We propose a different loss function that can 
be better estimated given the lack of data and incorporate
domain expert knowledge.

\vspace{-3pt}
\section{Setting and Background}\label{sec:setting}
\vspace{-3pt}
We consider an episodic stochastic
decision processes with a finite action space
$\mathcal{A}$, continuous state space $\mathcal{X} \in \mathbb{R}^M$, reward function
$R : \mathcal{X} \times \mathcal{A} \rightarrow [0, R_{max}]$ and discount factor
$\gamma \in [0, 1]$. A policy $\pi$ maps the state space to a probability distribution over 
the action space, and we assume each episode lasts at most $H$ steps. A set of trajectories  
$\mathcal{T}=\{\tau_1,\dots,\tau_N\}$ is provided. Each trajectory $\tau_i$ consists of a  state $x_t$, action $a_t$ and observed reward $r_t$ at step $t$, $\tau_i = \{x^i_0,a^i_0,r^i_0,\dots,x^i_H\}$. 
Actions are generated by following a known behaviour policy $\pi_b$, $a_t \sim \pi_b(s_t)$.  We denote the evaluation policy by $\pi_e$. 




\vspace{-3pt}
\section{Framework for Subgroup Identification}
\vspace{-3pt}

Our focus is to robustly quantify the expected benefit or cost of switching from a behavior policy to a proposed evaluation policy on subsets of the population. To do so it is helpful to extend the standard notion of the treatment effect to the (sequential decision) policy treatment effect. We define the individual treatment effect $t(x;\pi_e,\pi_b)$ for a possible initial state $x$ as
\begin{align}\label{eq:ind_t}
    t(x;\pi_e,\pi_b) = \mathbb{E}_{\pi_e} \left[\sum_{t=0}^H \gamma^t r_t|x_0=x \right] - \mathbb{E}_{\pi_b} \left[\sum_{t=0}^H \gamma^t r_t|x_0=x\right].
\end{align}
Before we introduce our definition of group treatment effects, we first define a partitioning
over the state space by $L=\{l_1, \dots, l_M\} \in \Pi$, such that $\bigcup\limits_{i=1}^M l_i = \mathcal{X}$ and
$\forall i, j: \quad l_i \bigcap l_j =\emptyset$. Define the partition function 
$l(x; L) = l_i$ such that $x \in l_i$. Given a partitioning $L$, 
partition-value function for a policy $\pi$ can be defined as:
\begin{align*}
 v(x; L,\pi) = \underset{{\substack{x'\sim\mathcal{X}\\ a\sim\pi(.|x')}}}{\mathbb{E}}\left[\sum_{t=0}^H \gamma^t r_t|x_0=x', x'\in l(x;L)\right]
\end{align*} 
Using this function we can define group treatment effect, similar
to the individual treatment effect as,
\begin{align}\label{eq:group_treatment_effect}
    T(x;L, \pi_b, \pi_e) = v(x;L, \pi_e) - v(x;L, \pi_b)
\end{align}
note that group treatment effect is constant within every $l_i$,
and we refer to each $l_i$ as a group. With little abuse of notation we denote the individual treatment
effect by $t(x)$ and group treatment effect by $T(x;L)$ and may interchangeably use group and subgroup.

\subsection{Group treatment effect estimator}
Given a partition $L$, a set of trajectories $\mathcal{T}$,
the behaviour policy $\pi_b$ and an evaluation policy $\pi_e$
the following estimator defines the group treatment effect
estimator for an initial state $x$ over a dataset $\mathcal{D}=\{(x_0, \rho_0, g_0), \dots, (x_N, \rho_N, g_N)\}$,
\begin{align}\label{eq:estimator}
    \hat{T}(x; L) = \frac{1}{\left| \{x_i|x_i \in l(x;L)\} \right|}\sum_{i|x_i \in l(x;L)} (\rho_i g_i - g_i)
\end{align}
Where, $x_i = x_0^i$ is the initial state of a trajectory $\tau_i$,
$g_i$ is the discounted return $g_i = \sum_{t=0}^H \gamma^t r^i_t$ and $\rho_i$ is the importance sampling ratio $\rho_i = \prod_{t=0}^H\frac{\pi_e(a^i_t|x^i_t)}{\pi_b(a^i_t|x^i_t)}$. It is
straightforward to show that $\hat{T}(x;L)$ is an unbiased
estimator of $T(x;L)$ in every group.

Following much of the literature \citep{athey2016recursive, thomas2016ope} 
we focus on the MSE criteria to rank different
estimators defined by different partitioning; however, as explained
later, we modify this loss in multiple ways to account
for our goal. 
\begin{align*}\label{eq:mse}
    MSE(\hat{T}; L) = \underset{x\sim\mathcal{X}}{\mathbb{E}} \left[\left(t(x) - \hat{T}(x; L)\right)^2\right]
\end{align*}

Note that MSE loss is infeasible to compute, as we 
do not observe treatment effect $t(x)$.  However, we show that it is equivalent to an expectation over quantities that can be estimated from data.

\begin{theorem}\label{th:loss}
    For a given partition $L\in \Pi$, let $T(x;L)$ be the group treatment
    effect defined in equation \ref{eq:group_treatment_effect}, $t(x)$ be the individual treatment effect as defined in equation \ref{eq:ind_t} and $\hat{T}(x;L)$ an unbiased estimator of
    $T(x;L)$. The following display impose the same ranking over the partitions as the MSE loss in equation \ref{eq:mse}:
    \begin{align}
         -\mathbb{E}_{x\sim\mathcal{X}}\left[\hat{T}^2(x;L)\right] + 2~\mathbb{E}_{x\sim\mathcal{X}}\left[\mathbb{V}\left[\hat{T}(x;L) \right]\right]
    \end{align}
    Where $\mathbb{V}[\hat{T}(x;L)]$ is the variance of the estimator $\hat{T}(x;L)$. 
  
\end{theorem}
The proof is provided in the supplementary materials. 
The results of theorem \ref{th:loss} suggest an estimatable quantity that can be used to select among different potential partitions. More precisely, given a dataset $\mathcal{D}$ the empirical adjusted MSE can be written as,
\begin{align}\label{eq:emse}
    EMS&E(\hat{T}; L) = -\frac{1}{N}\sum_{i=1}^N \hat{T}^2(x_i;L) + \frac{2}{N}\sum_{i=1}^N \mathbb{V}\left[\hat{T}(x_i;L) \right]
\end{align}
Where $\mathbb{V}[\hat{T}(x_i;L)]$ is the variance
of the estimator $\hat{T}(x_i;L)$ in the subgroup $l_i$ s.t. $l_i=l(x_i;L)$. We next describe an algorithm to construct a good partition that minimizes the above loss, as well as how we alter the above loss to advance our goal of being able to robustly estimate group treatment effects.

\vspace{-3pt}
\section{A Practical and Effective Algorithm for Subgroup Identification}\label{sec:group}
\vspace{-3pt}

In this section we first assume access 
to a loss function $\mathcal{L}(L)$ and
describe the recursive partitioning
algorithm to minimize it. Further we 
discuss the modifications we apply
to the empirical adjusted MSE in section \ref{sec:loss} to
obtain the loss function $\mathcal{L}(L)$.

\subsection{Algorithm}
In order to partition the feature space to different subgroups we
minimize a loss function $\mathcal{L}(L)$ with recursive
partitioning, $\min_{L \in \Pi}~ \mathcal{L}(L)$. First
in the partitioning phase, similar
to classification and regression tree (CART)
\citep{breiman1984classification}, we build a tree by
greedily splitting the feature space to minimize the
loss function, and stop splitting further when there is no
such split that results in the reduction of the 
loss function (partitioning phase), we call 
this a treatment effect tree.

After building the treatment effect tree, each leaf $l_i$ is a group
and we can form an estimate of the group treatment effect by
equation ~\ref{eq:estimator} (estimation phase). Note that in
the estimation phase, different OPE methods
such as model based and doubly robust \citep{thomas2016ope, liu2018representation} can be used to form the prediction. 
In this work, we use the same estimator in the partitioning
and estimation phase and mainly focus on
developing a loss function to be used in the partitioning phase.
Additionally, we compute confidence
intervals around our estimation by bootstrapping. 

\subsection{Loss Function}\label{sec:loss}

One way to estimate the empirical adjusted MSE in equation \ref{eq:emse}
is by substituting the variance term with the sample variance
of the  estimator. This is similar to the loss proposed by \citet{athey2016recursive} in the non-sequential setting. However, estimation of the sample variance may be very noisy due to limited data, particularly in our sequential setting. 
A mis-estimation of the variance may result in an avoidable undesirable  split in partitioning phase that would have not happened given a better estimate of the variance. Indeed, over-splitting is a common failure mode of using this loss function as we demonstrate in  our experiments. 



\paragraph{Variance Estimation.} To mitigate
the issue of over-splitting 
we modify the loss function by a proxy of 
the variance term which can be computed efficiently. First we show that the variance of the treatment effect estimator can be upper bounded by quantities that can be easily computed from data.
\begin{theorem}\label{th:variance}
Given a dataset $\mathcal{D}=\{(x_0, \rho_0, g_0),\dots,(x_N,\rho_N,g_N)\}$ and the treatment effect estimator defined by $\hat{T} = \frac{1}{N}\sum_i(\rho_i - 1)g_i$. The variance of $\hat{T}$ satisfies the following inequality,
\begin{align}\label{eq:upp_bound}
    \mathbb{V}[\hat{T}] \leq \left\lVert g\right\lVert^2_\infty\left(\frac{1}{ESS} - \frac{1}{N}\right)
\end{align}
\noindent where, $ESS$ is the effective sample size.
\end{theorem}

Note that in the special case of behaviour policy being the same 
as the evaluation policy, this bound evaluates to zero. We denote the RHS of equation \ref{eq:upp_bound} by
$\mathbb{V}_u[.]$. In our work, we use $\mathbb{V}_u[.]$ in each leaf as a proxy of variance of 
the estimator in the leaf. That is,
\begin{align}\label{eq:upper_bound}
    \mathbb{V}_u[\hat{T}(x_i;L)] = ||g(x)||^2_\infty\left(\frac{1}{ESS(l_i)} - 
         \frac{1}{|l_i|} \right), 
\end{align}
where we use the common ESS estimate $ESS(l_i)$ by $\widehat{ESS}(l_i) = \frac{(\sum_j \rho_j)^2}{\sum_j \rho_j^2}$
where the sum is over samples inside the group $i$, $\{j|x_j \in l_i\}$~\citep{mcbook}. $\mathbb{V}_u[.]$ can be
computed efficiently and the conservative variance estimation using $\mathbb{V}_u[.]$ avoids the problem of
variance underestimation. 
Note that another approach to get a better estimate of the variance could be to leverage bootstrapping; however, using such a procedure is not feasible in the partitioning phase 
due to it's prohibitive high computational cost, since it has to be done for every 
evaluation of the loss function.

\paragraph{Regularization} In many applications, actionable
information needs to satisfy certain conditions. For example,
a clinician may consider a knowledge of group treatment effect
useful, if we can guarantee with high probability that the 
treatment effect is $\alpha$ bounded away from zero. 
Our above loss function which is focused on minimizing 
the mean squared error would not
necessarily identify these practically relevant subgroups. 


Therefore we now 
introduce a regularization term into our loss function to encourage finding such partitions where some subgroups have treatment effects that are bounded away from zero. In order to do so we use Cantelli's inequality to derive a lower
bound on the estimator defined in equation \ref{eq:estimator}. While this is a weaker bound than Bernstein, this allows us to avoid assuming we have access to an upper bound on the importance weights. 
We do assume that the function $\hat{T}[x;L]: \mathcal{X} \rightarrow \mathbb{R}$, is a bounded function ($||\hat{T}(x;L)||_\infty < \infty$). We start by writing Cantelli's inequality
applied to the random variable $\hat{T}(x;L)$,
\begin{align*}
    \mathbb{P}\left(\hat{T}(x;L) - \mathbb{E}[\hat{T}(x;L)] \geq \lambda\right) \leq \frac{1}{1+\frac{\lambda^2}{\mathbb{V}[\hat{T}(x;L)]}}
\end{align*}

 Assigning $\delta$ to the right hand side and
 considering the complementary event,
we have with probability $1-\delta$,
\begin{equation}\label{eq:cant}
        \mathbb{E}[\hat{T}(x;L)] \geq \hat{T}(x;L) - \sqrt{\frac{1-\delta}{\delta} \mathbb{V}[\hat{T}(x;L)]}
\end{equation}
With Equation \ref{eq:cant} we define the (margin $\alpha$) regularization term 
\begin{align}\label{eq:reg}
     \mathcal{R} (x_i;L, \alpha) = \max \left\{0, \alpha - \left(|\hat{T}(x_i;L)|  - c \sqrt{\mathbb{V}_u[\hat{T}(x_i;L)]}\right) \right\}
\end{align}

Note that we used $\mathbb{V}_u[.]$ instead of $\mathbb{V}[.]$ in
equation \ref{eq:reg} to avoid issues arising from under estimation
of the variance. Although we can obtain $c$ by setting a specific
value of $\delta$, we view this as a tuning parameter for
regularization. 

In a similar fashion, other types of regularizations depending on
domain expert's input can be used in partitioning phase. For example, a domain expert may be interested to take
more risks if the predicted group treatment effect is larger. This type of
information can be incorporated as,
\begin{align*}
    \mathcal{R}(x_i;L, \alpha) = \max\left\{0, \alpha - \frac{|\hat{T}(x_i;L)|}{c \sqrt{\mathbb{V}_u[\hat{T}(x_i;L)]}} \right\}
\end{align*}

\paragraph{Loss Function} By combining the regularization term
and using the proxy variance, we obtain our final loss function.
\begin{align}\label{eq:loss}
    \nonumber \mathcal{L}(L) = -\frac{1}{N}\sum_{i=1}^N \hat{T}^2(x_i;L) &+ \frac{2}{N}\sum_{i=1}^N \mathbb{V}_u\left[\hat{T}(x_i;L) \right]+\frac{C}{N} \sum_{i=1}^N \mathcal{R}(x_i;L,\alpha), 
\end{align}
where $C$ is the regularization constant. This loss is minimized using recursive partitioning. We call our algorithm GIOPE, group identification in off-policy policy evaluation. Note in Theorem \ref{th:loss}
we relied on $\hat{T}(x;L)$ be an unbiased estimate of $T(x;L)$. To accomplish this with our chosen estimator for $T(x;l)$ we use  independent set of samples for
partitioning phase and the estimation phase. 
The importance of sample splitting 
to avoid overfitting during off policy estimation is well studied (e.g. \citep{craig2020finding,athey2016recursive}). 

\section{Experiments}\label{sec:experiments}

We illustrate how our approach allows us to partition
the feature space into subgroups such that we can
make confident and
accurate predictions of the group treatment effect.
We empirically
evaluate our method in sequential decision making settings,
compare to the baseline and perform ablation 
analysis to show the benefit of each modification we have proposed.

We start by a simple toy example to illustrate the 
benefits of our method, later we evaluate our method on a simulated health care example, management of sepsis
patients~\citep{oberst2019counterfactual}. Additionally, we use freely available MIMIC III dataset 
of ICU patients \citep{johnson2016mimic}, and focus on
the sepsis cohort \citep{komorowski2018artificial} to show how our method can be used with real world data.
We provide the code for all experiments in the supplementary
materials. We compare
to causal forests (CF) \citep{athey2019generalized} that was developed
for non-sequential setting. To the best of our knowledge, CF is one the best performing algorithm in non-sequential setting that yields a good performance across different domains.

\subsection{Simple Illustration} 

We consider a 1 dimensional toy MDP to illustrate the difference between
our method and methods developed for non-sequential setting. The toy MDP 
has the transition dynamics $x_{t+1} = clip(x_t + \kappa \times a_t + \epsilon, 0, 1)$, where the function $clip(x, a, b)$, clips the value of
$x$ between $a$ and $b$ and reward function $r(x) = 1-|x-0.5|$. Details of 
the experiment can be found in the supplementary materials. 

\begin{figure}[tb]
    \begin{tabular}{cccc}
    \hspace{-20pt}
    \includegraphics[width=0.25\linewidth]{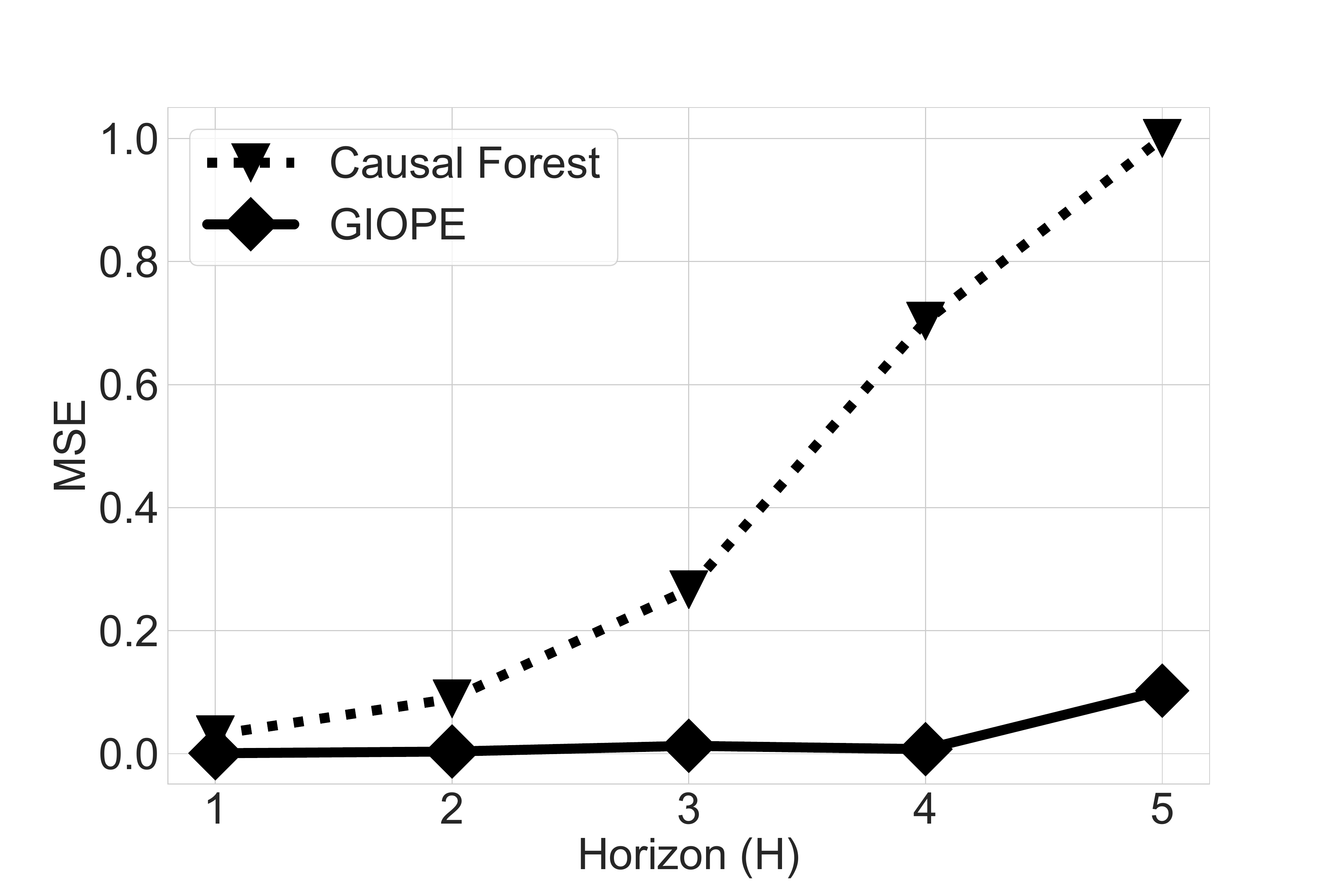} &
    \includegraphics[width=0.25\linewidth]{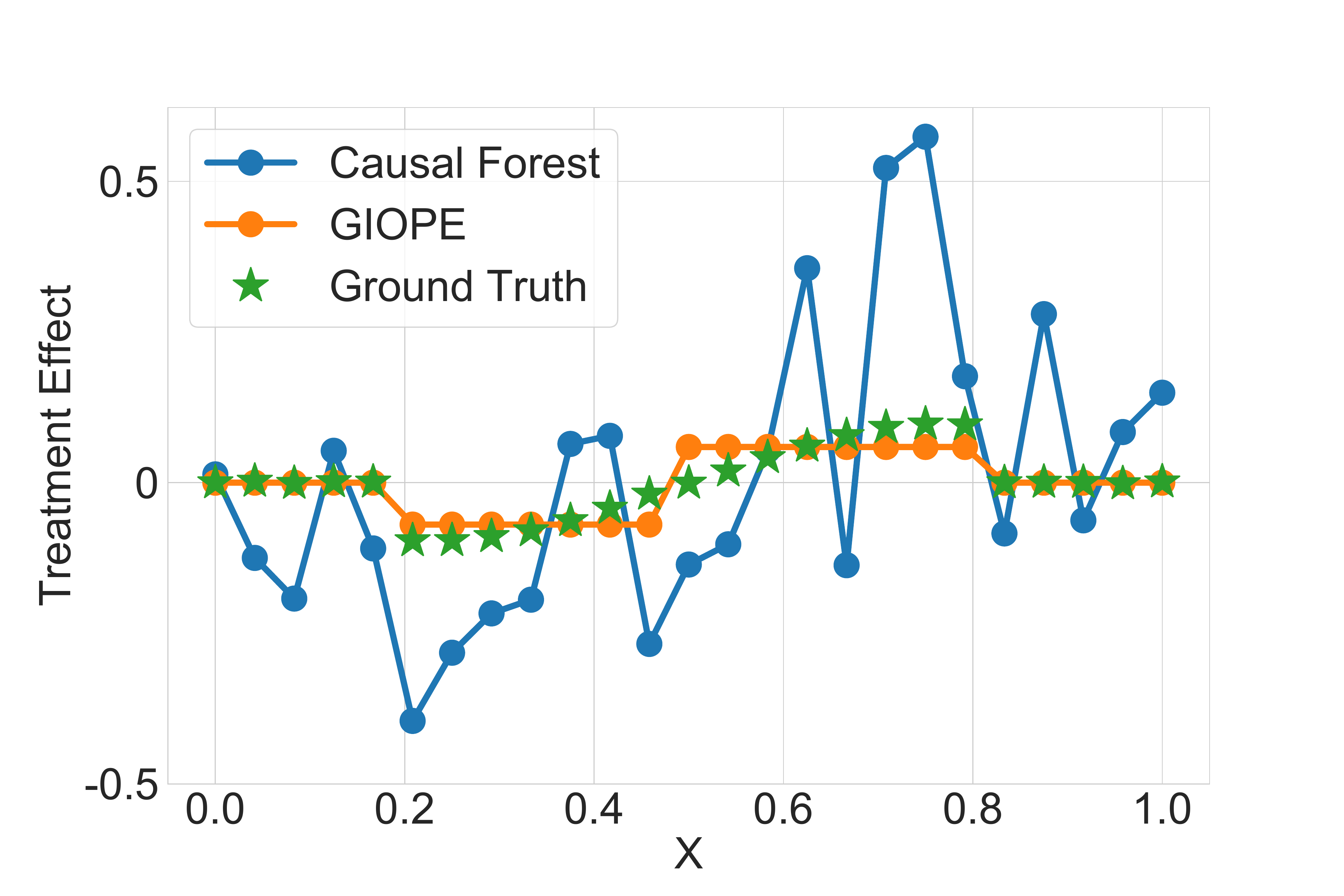}&
    \includegraphics[width=0.23\linewidth]{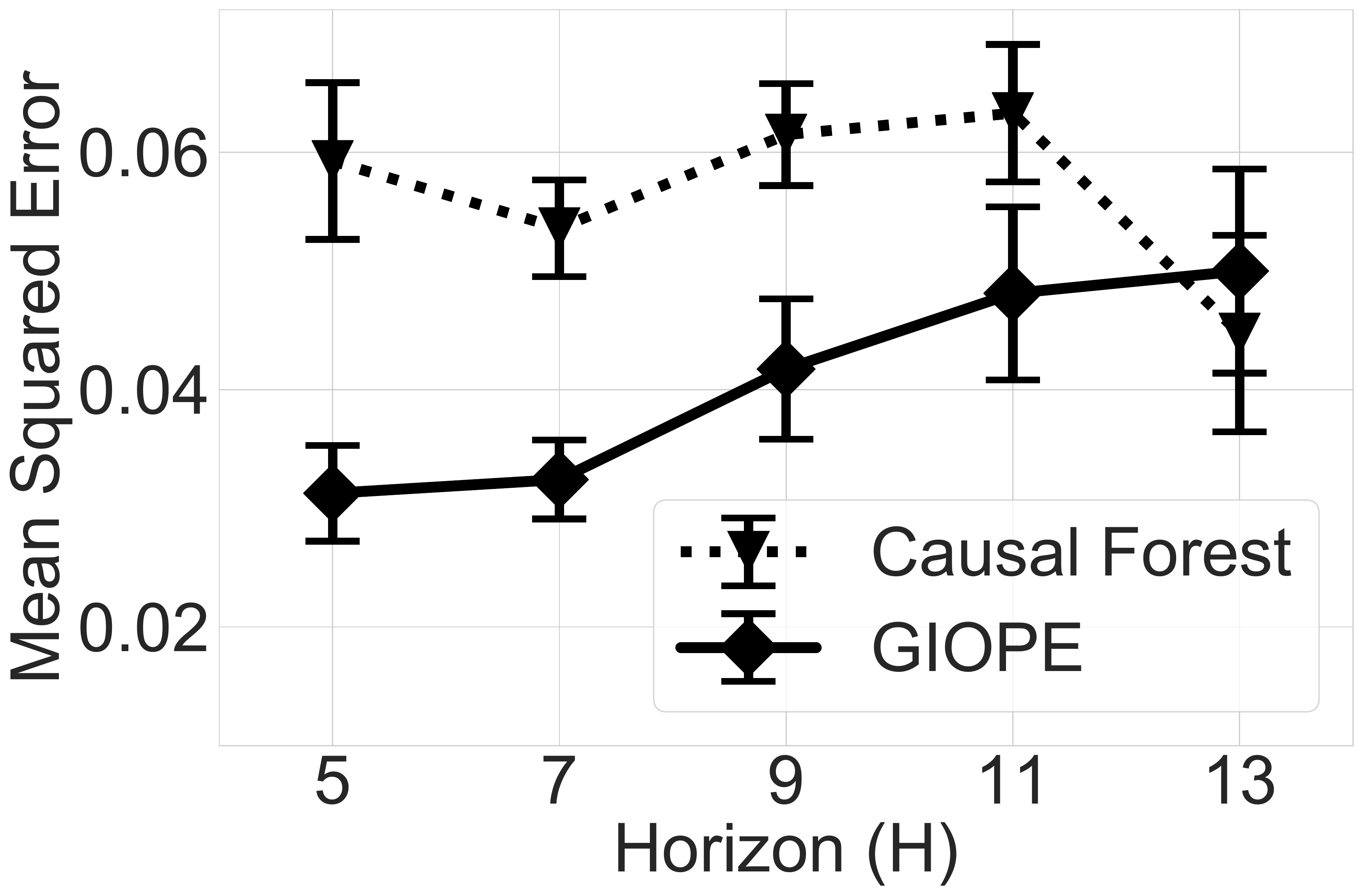}&
    \includegraphics[width=0.23\linewidth]{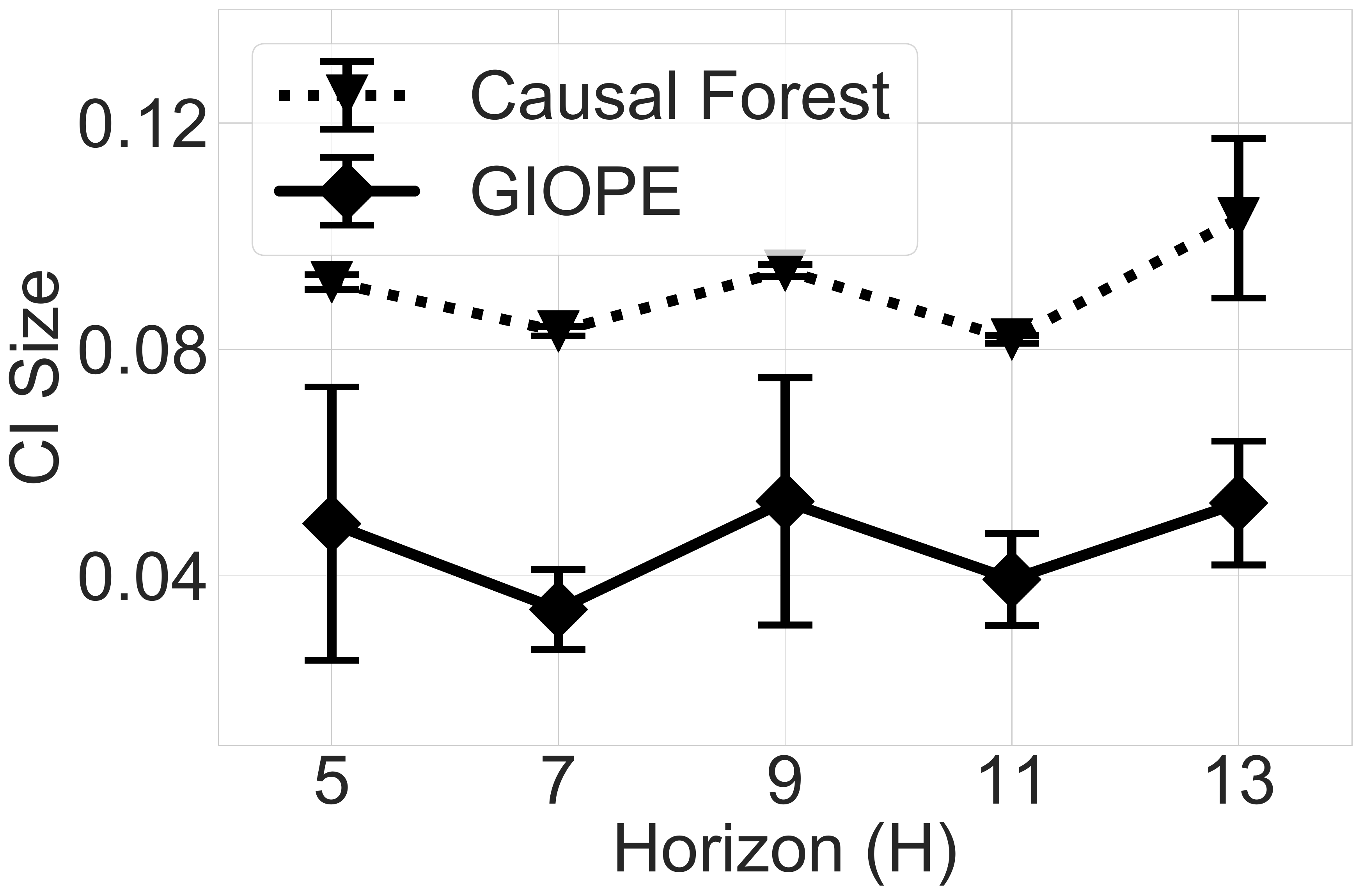}\\
    (a) &  (b) & (c) & (d)
\end{tabular}
\caption{\textbf{Toy MDP:} (a) Mean squared error of treatment effect prediction for our method and causal forest(CF). (b) True and predicted treatment effect for different values of $x$ for our method and causal forest. \textbf{Sepsis simulator:} comparison with causal forest (CF).
(c) Mean squared error of prediction. (d) Average size of the 95\% confidence intervals (CI}\label{fig:res}
\end{figure}

We look at the mean squared error of the 
treatment effect prediction on 25 equally spaced points in $[0, 1]$. Figure \ref{fig:res} (a) compares the MSE
between our method with causal forest (CF).
GIOPE shows smaller MSE and as the horizon increases
the benefit is more apparent. Figure \ref{fig:res} (b)
shows the predicted
value of the treatment effect for our method and causal 
forest for horizon $H=4$ along with the true treatment effect for different values of $x$. This illustrates the reason
of performance gap. Our
method partitions the
state space and makes the same prediction for each subgroup that results in more accurate predictions, whereas causal forests over-splits and compute different
values of the treatment effect for every value of $x$ which are often inaccurate.

\subsection{Sepsis Simulation}

There has been growing number of literature that seek
to learn an automated policy to manage septic patients in the
ICU,  reader may find a short review in
~\citet{gottesman2019guidelines}. However, newly suggested
decision policies may be beneficial for some subgroups of patients
while harmful to others. We use the sepsis simulator developed 
in \citet{oberst2019counterfactual} to show case this scenario
and evaluate our models in detecting such subgroups. 

\paragraph{Simulator} In this simulator
each patient is described by four vital signs \{heart rate, blood
pressure, oxygen concentration
and glucose level\} and a binary indicator of diabetes, that take
values in a subset of \{very high, high, normal, low, very low\}, that
results in a state space of size $|S| = 1440$. In each step
the agent can take an action to put the patient on or off of
treatment options, \{antibiotics, vasopressors, and mechanical
ventilation\}, so that the action space has cardinality $|A|=2^3$.
Each episode run until the horizon $H$ which incurs the 
reward of -1 upon death, +1 upon discharge and 0 otherwise. We use a discount factor of $\gamma = 0.99$ across all experiments, and all
reported results are averaged over 15 different runs.

\paragraph{Data Generation} In order to form the behaviour and the evaluation policy we
assume both policies act nearly optimal with some modifications.
We perform policy iteration to find the (deterministic) optimal
policy for this environment and soften the policy by subtracting
$0.1$ probability from the optimal action and equally
distributing it among other actions, we call this policy $\pi_{st}$. We assume the behaviour
policy $\pi_b$ is similar to $\pi_{st}$ except it has
has $15\%$ less chance of using the mechanical ventilator.
On the other hand, the evaluation policy $\pi_e$ is similar
to $\pi_{st}$ but has $20\%$ less chance of using the vasopressor. Notice that, the evaluation policy utilized
the mechanical ventilator more and vassopressor less
than the behaviour policy. Regardless of the horizon,
the evaluation policy achieves better expected discounted
return than the behaviour policy. However, there are
subgroups of individuals, for example diabetics, that will worse off by using the evaluation
policy. We generate $50000$ trajectories using the behaviour policy.


\paragraph{Comparison} First we look at the mean squared
error computed on the individual level.
In order to compute this value, for each individual in the test set
that consists of $n=20000$ samples from the same distribution as the training
set, we sample $30$ different trajectories using the evaluation policy and the behaviour 
policy to compute the true treatment effect for each individual. 
Figure \ref{fig:res} shows the mean squared error of
prediction made by our method versus causal forest (CF). As shown in Figure \ref{fig:res} (c), our method outperforms the baseline but as horizon 
increases both models struggles to generate valid results. Panel (d)
of Figure \ref{fig:res} shows the average size of the 95\% confidence
intervals. This highlights one of the main benefits
of our method, that is more accurate prediction along
with tighter confidence intervals.

\paragraph{Identified Subgroups} Our methods can identify subgroups with significant negative treatment effect, those groups represent
patients with diabetes and elevated heart rate. For detailed description of each
subgroups for different horizons please refer to
supplementary materials. This qualitative analysis
unfortunately cannot be done for methods like causal
forest as they are not designed to yield distinctive
subgroups.


\begin{figure}[tb]
    \centering
\begin{tabular}{cccc}
    \includegraphics[width=0.23\linewidth]{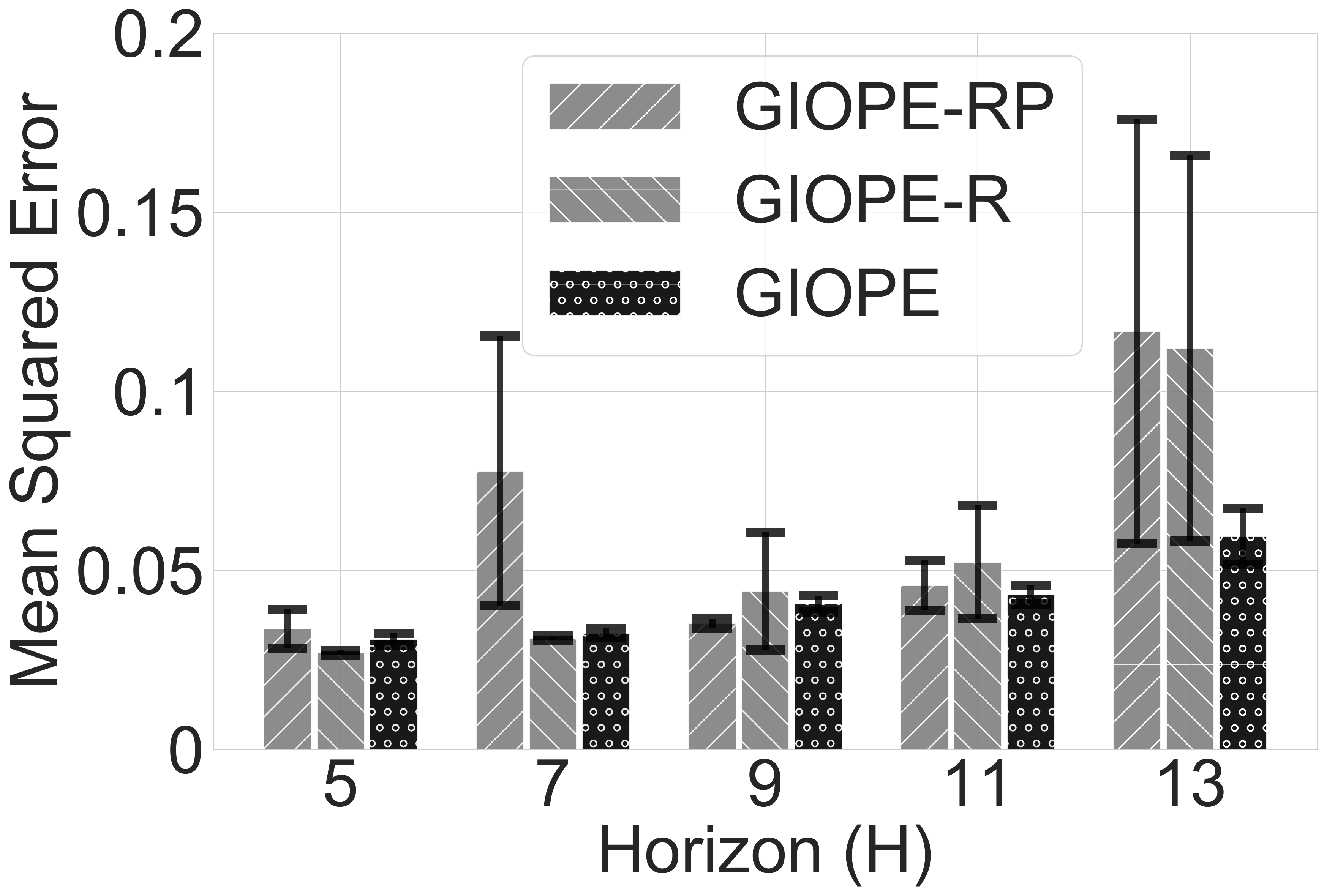} &
    \includegraphics[width=0.23\linewidth]{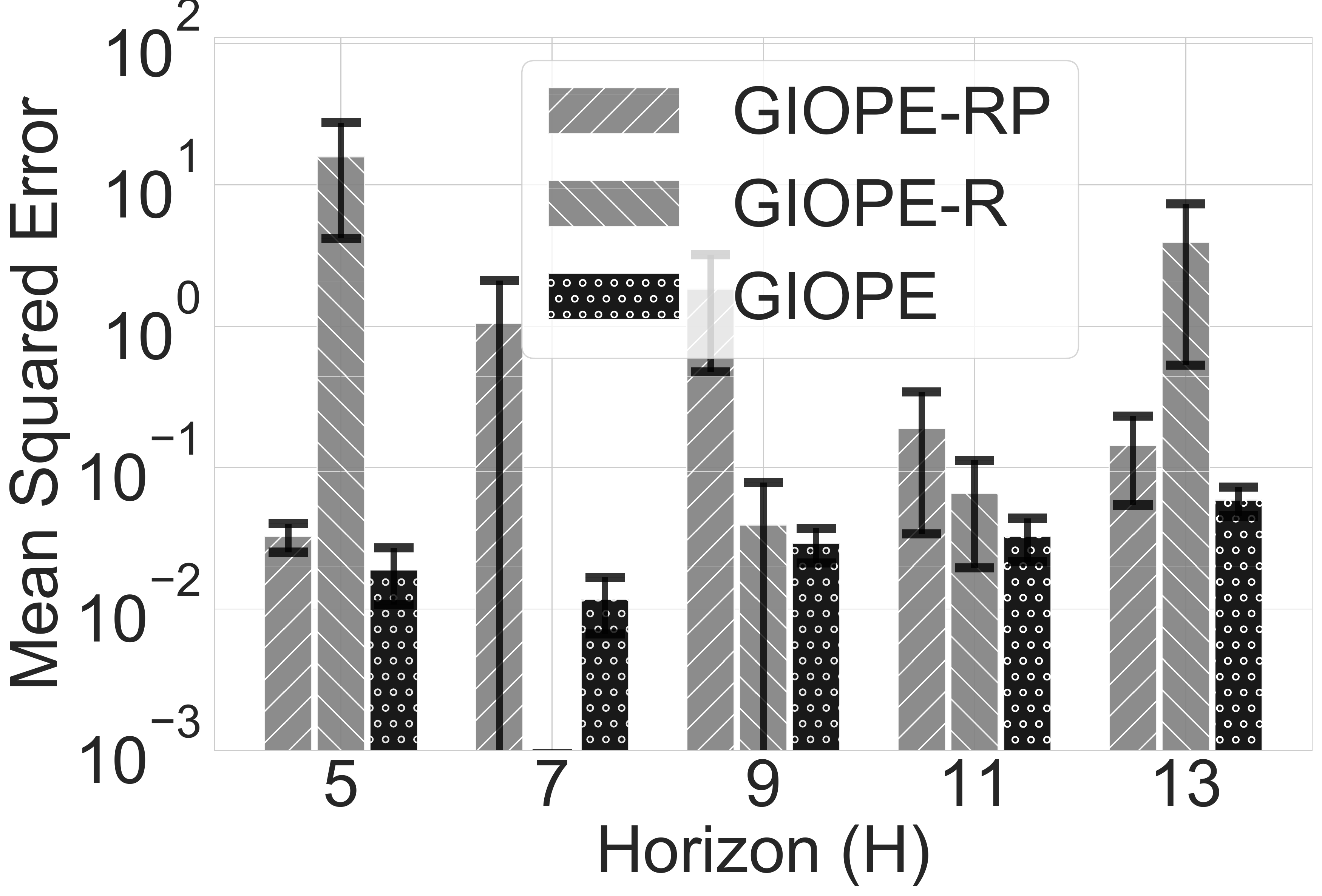} &
    \includegraphics[width=0.23\linewidth]{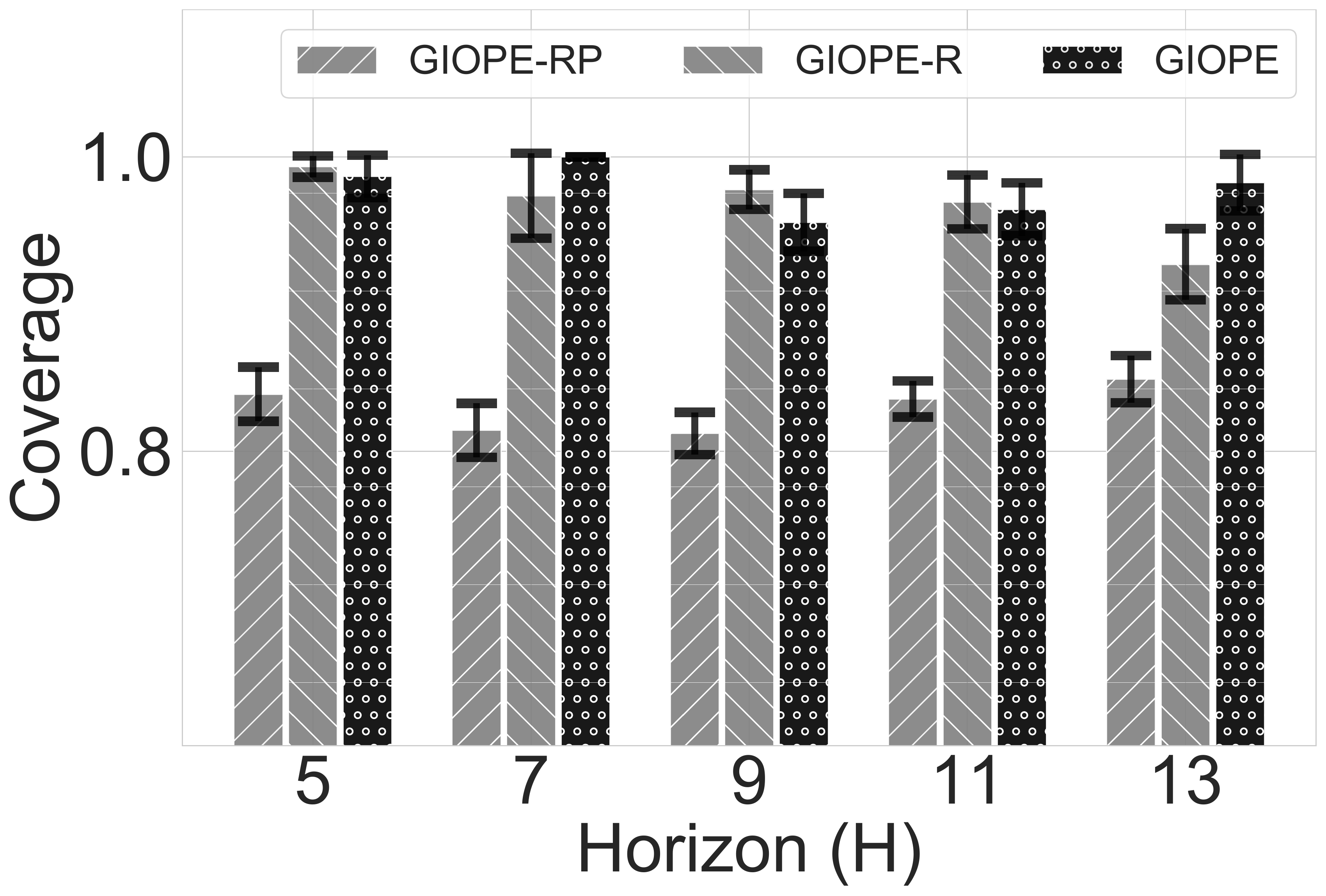} & \includegraphics[width=0.23\linewidth]{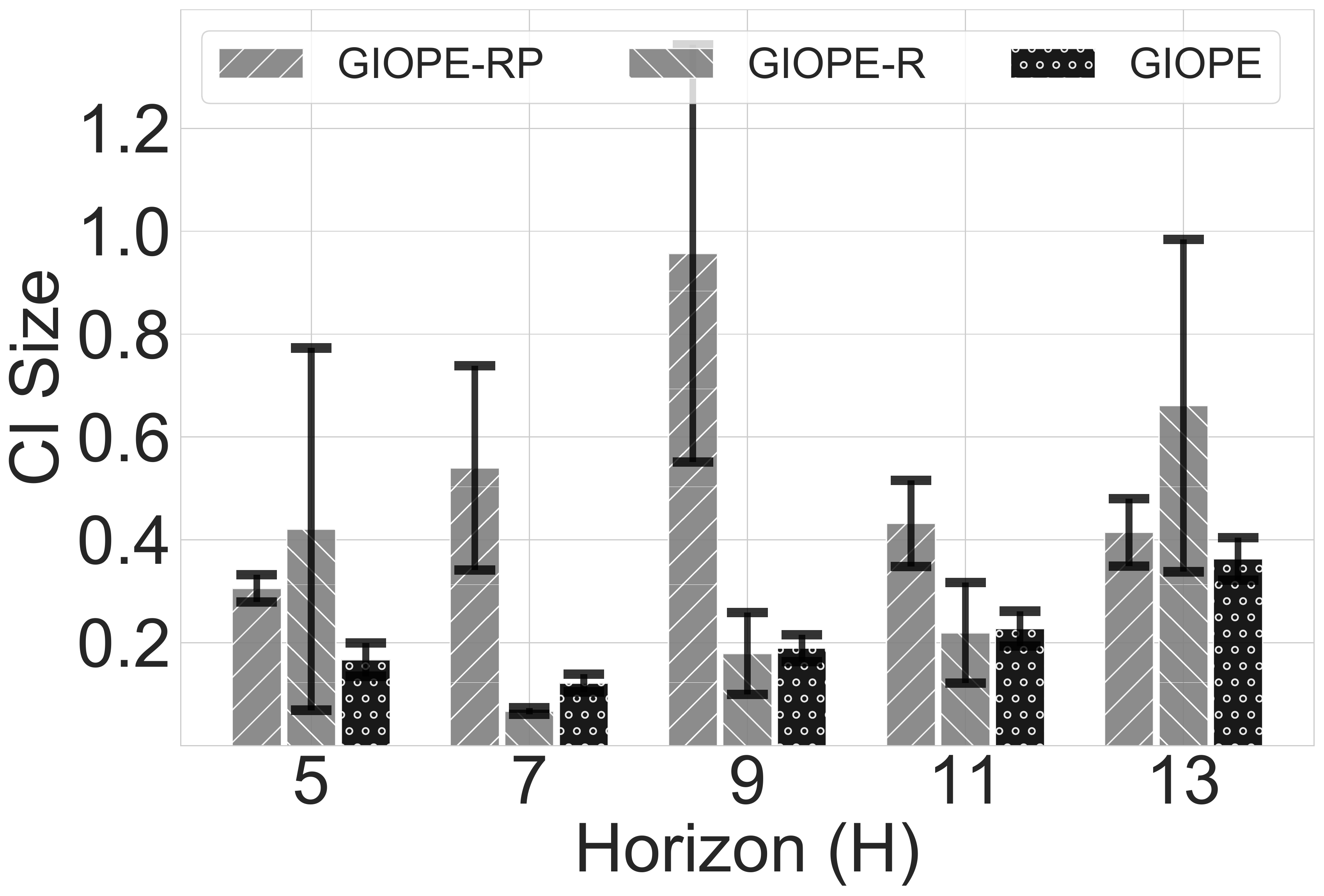} \\
    (a) & (b) & (c) & (d)\\
\end{tabular}
    \caption{Ablation study. (a) Mean squared error 
    computed on individual level. (b) Group mean squared error. Coverage: (c) Percentage of groups that the true group treatment effect is
    covered by the 95\% confidence interval. (d) Average size of confidence
    intervals.}
    \label{fig:ablation}
\end{figure}


\paragraph{Ablation Study} In order to showcase the benefit
of each modification that we proposed, we perform ablation
study on the sepsis simulator. We compare three different
methods with 15 different runs.
\begin{enumerate}
    \item \textit{GIOPE}: Using the loss function presented in equation \ref{eq:loss}. In all experiments, the value of regularization is set to $C=5.0$ and the margin $\alpha = 0.05$. We found that changing this regularization value has little effect on the results as we presented in the Appendix.

    \item \textit{GIOPE - Regularization (GIOPE-R)}: Using the
    loss function in equation \ref{eq:emse}
    with the suggested proxy variance in equation
    \ref{eq:upper_bound}.
    
    \item \textit{GIOPE - Regularization and Proxy Variance (GIOPE-RP)}: This method uses the
        loss function presented in equation \ref{eq:emse} with
        the sample variance estimate. Note that this basic version is similar to the loss
    function proposed in \citet{athey2016recursive}.
\end{enumerate}


First we look at mean squared error computed
on the individual level. As shown in Figure \ref{fig:ablation} (a) our method shows significant
benefits compared to GIOPE-R and GIOPE-RP. This comes with an important observation that our method also shows more stability as the performance does not fluctuates as much
across different horizons as well as having
smaller standard errors. 
Note that, our method
do not optimize for this objective and the individual
mean squared error is best minimized
with the sample variance in the limit of infinite
data, the benefit comes as
an externality of avoiding to predict each individual separately.

Next we look at mean squared error in group treatment effect. That is, for
a groups $i$, denote the prediction of the group treatment effect by $\hat{g}_i$ and the true group treatment effect by $g_i$, then the group MSE is defined
as $\frac{1}{G}\sum_{i=1}^G (g_i - \hat{g}_i)^2$, where $G$ is the total
number of groups. Figure \ref{fig:ablation} (b)
shows the MSE in group treatment effect as we increase the horizon. Similar to individual MSE, our method
obtains lower MSE and displays more stability across
different horizons. 
This stability is mainly due to avoiding to over split.
For example, average number of discovered groups in
GIOPE-RP method for horizon 13 is 26 whereas for other GIOPE-R is 5 and GIOPE is 4. 

Finally we look at coverage. Figure \ref{fig:ablation}, panel (c) shows the coverage of 95\%
confidence intervals of the true group treatment effect for
different methods and horizons. Methods that use variance
proxy instead of sample variance show consistently more
coverage. Figure \ref{fig:ablation} panel (d)
shows the average size of the confidence interval for each group
treatment effect prediction. 
This indicates that using the upper bound along with regularization (GIOPE)
yields more coverage while offering tighter confidence intervals. 
This observation highlights the main benefit of using
regularization along with proxy variance that allows us 
to discover groups that we can more accurately and confidently
predict their treatment effect. Smaller standard error of
confidence intervals size highlights 
the stability of GIOPE across different runs.

\subsection{ICU data - MIMIC III}
To show how our method can be used on a real data set, we use
a cohort of 14971 septic patients in the freely accessible MIMIC III dataset \citep{johnson2016mimic}. Prior work~\cite{komorowski2018artificial} used off policy learning and proposed a new decision policy that might provide improved patient outcomes on average.  More details on the experiment setup and the off policy learning approach used are in the supplementary materials. Using weighted 
importance sampling the estimated value of the decision policy is $65.33$
with effective sample size of $146.8$ which
suggest an increase of $2.43$ on the survival chance compared to the behaviour
policy. Here we take this decision policy and estimate its impact on different potential subgroups.

\begin{figure}[tb]
    \centering
    \begin{tabular}{cccc}
    \includegraphics[width=0.23\linewidth]{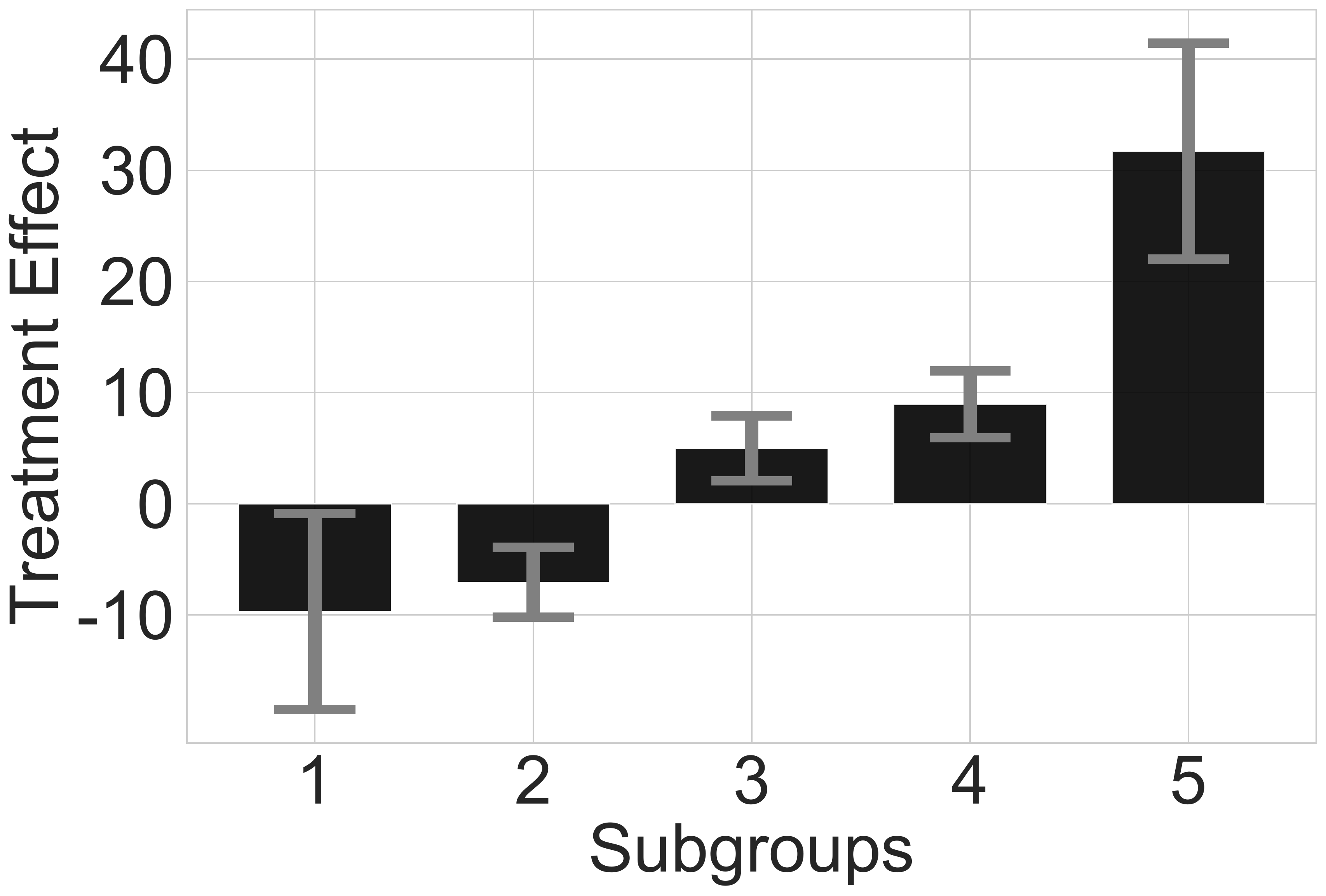} & 
    \includegraphics[width=0.23\linewidth]{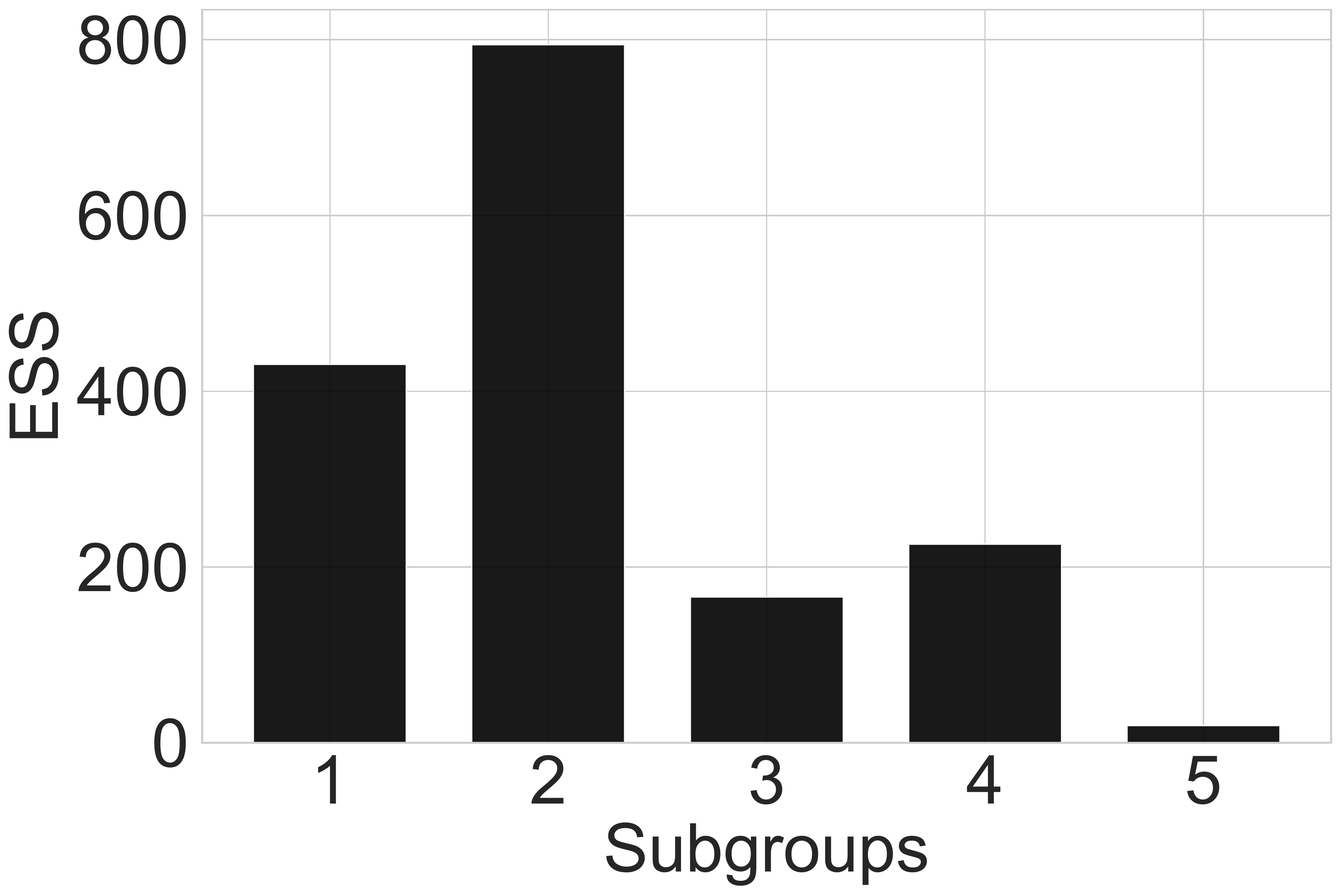} &
    \includegraphics[width=0.23\linewidth]{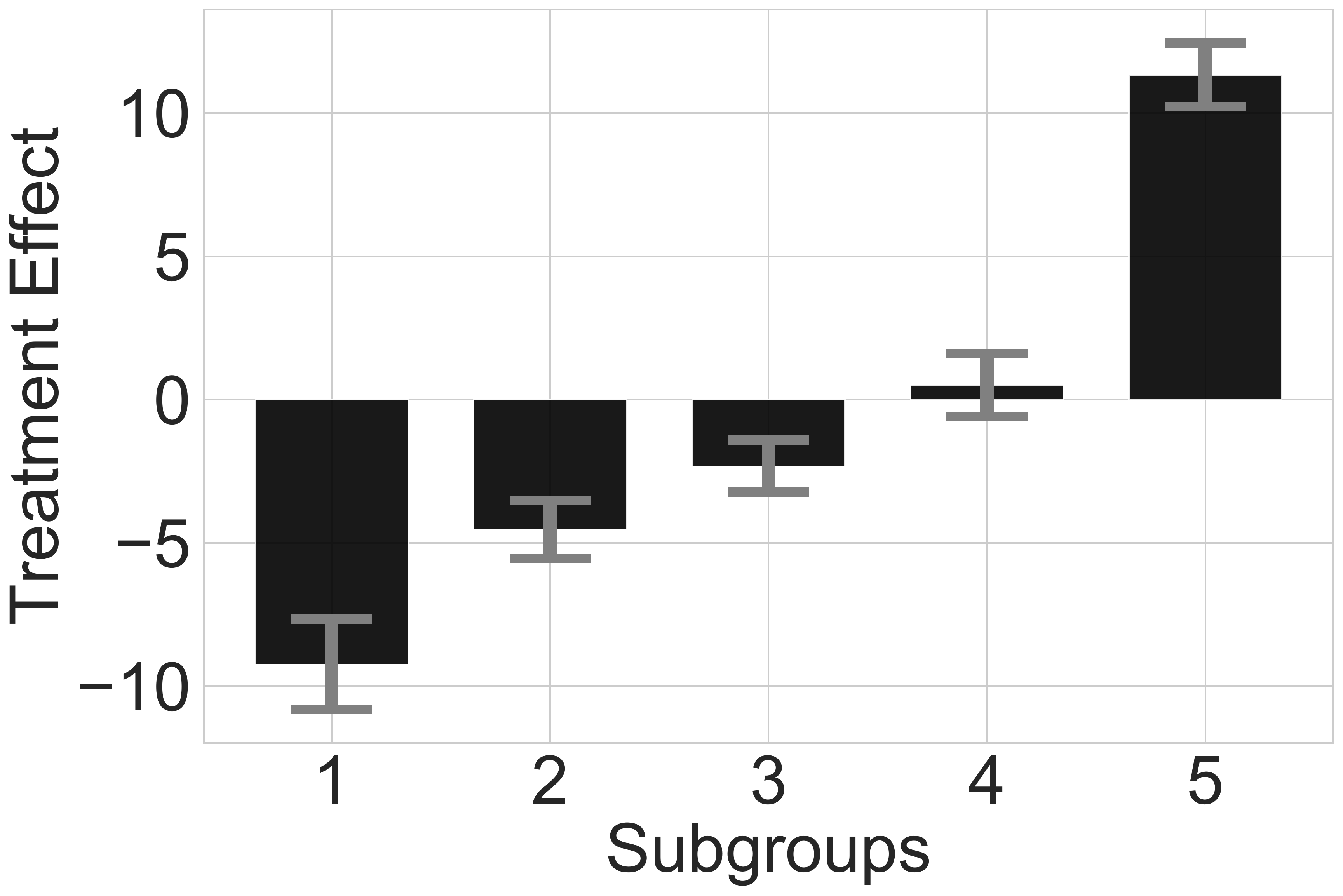} &
    \includegraphics[width=0.23\linewidth]{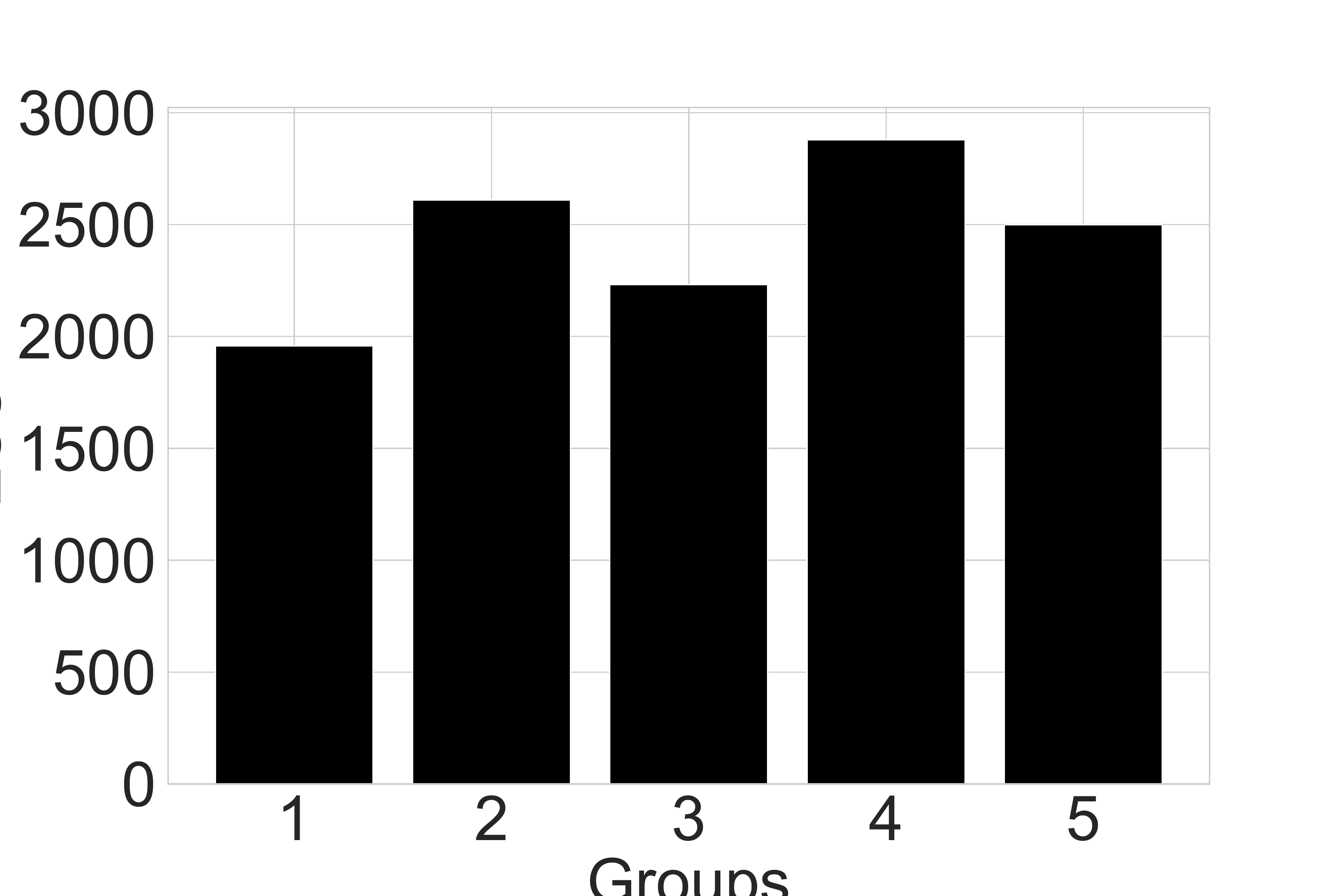}\\
    (a) & (b) & (c) & (d)
    \end{tabular}
    
    \caption{MIMIC III dataset. AI Clinician: Although positive treatment effect is predicted by weighted importance sampling on the full cohort, groups 1 and 2 will like be harmed by the evaluation policy. (a) Estimated treatment effect for each subgroup, (b) Effective sample size of weighted importance sampling for each subgroup. More aggressive vasopressor use: (c) Estimated treatment effect for each subgroup, (d) Effective sample size of weighted importance sampling for each subgroup}
    \label{fig:mimic}
\end{figure}

In  Figure \ref{fig:mimic} we present the five groups produced by our algorithm along with
their estimated group treatment effect (which is the difference between the baseline clinician policy and the decision policy) and
effective sample size of weighted importance sampling in each subgroup. While some of patients fall into subgroup 3, 4 and 5, there are a number of patients that may experience no benefit or even a potential negative treatment effect from the proposed new treatment policy (groups 2 and 1). 
This highlights how our method may be useful in identifying subgroups in which a new decision policy may be beneficial or harmful relative to the standard approach. 

We caveat the results in this section by noting that using IS based methods on real world datasets, and the MIMIC III dataset in particular is very susceptible to noise induced by the small effective sample size of the cohort \citep{gottesman2018evaluating}. Furthermore, our method is susceptible to this source of noise twice, as IS based estimators are used both in the partitioning phase and the estimation phase. However, despite their high susceptibility to noise, IS methods are often applied to the MIMIC III dataset for their theoretical properties, but their results for real data should be interpreted with caution. In our experiment we intentionally designed the decision policy close to the behaviour policy to avoid issues arising from small effective sample size. 

\paragraph{More Aggressive Use of Vasopressor} Additionally, 
we evaluated our method on a policy that utilizes vasopressor
more often than the behaviour policy. We estimate the behaviour
policy using KNN, and evaluate a policy that has 10\% more probability
mass on using vasopressor than the behaviour policy. Figure \ref{fig:mimic} (c) shows the subgroups found using our method along with the effective sample size of every group in display (d).

Group 1, 2, and 3 all show a negative treatment effect. Interestingly, these three groups have \texttt{SOFA} $> 1$ which indicates these patients are at high risk. Given the discussions we had with an intensivist, this is in agreement with their expectation that healthier patients are less likely to be harmed by more aggressive use of vasopressor, and sicker patients may be more at risk. This also highlights one of the main benefits of our method: it can be used to provide interpretable subgroups with differing potential treatment effects that may be used to support communication with clinicians around potentially beneficial alternate treatments, and who they might benefit. 
\section{Conclusion and Future Works}\label{sec:conc}
In this paper, we proposed a novel method to partition the feature space, enabling us to find subgroups that we can accurately and confidently predict the group treatment effect for them. 
Our approach is in contrast with previous methods that estimate individual-level treatment effects, yielding uncertain and less accurate predictions. We do so, by proposing a novel loss function that utilizes; 1. A proxy on the variance estimator that is easy to compute and stable; 2. A regularization term that incentivizes the discovery of groups with significant treatment effect and allows us to integrate domain expert's input into partitioning algorithm. We further evaluate our method on both simulated domains and real-world data. 

Our method can leverage the existing data to raise caution when necessary about a possible negative effect of the newly suggested decision policy on some subgroups. Additionally, results from our
method, when applied to observational data, can help to design multi-stage randomized trials that are powered toward detecting harm or benefit of the evaluation policy compared to the baseline policy for specific subgroups. However, it is crucial not to overly rely on the results of our method to justify certain treatments. Most of the findings, especially in health care, have to be validated via a randomized clinical trial or with an extensive discussion with the medical team.

Here, we discuss some limitations and avenues of future works. The current method only employs importance sampling to obtain an estimate of the treatment effect. However, it is well known that the importance sampling method can have high variance. Integrating other methods of off-policy policy evaluation in the partitioning phase, for example doubly robust or model-based method \citep{thomas2016ope}, can mitigate this issue and result in better partitioning. 

Additionally, we suggested using recursive partitioning to minimize the loss function; however, these greedy algorithms may fail to find the optimal solution; thus, an immediate question is how can we apply better optimization techniques.  Finally, we consider subgroups for the initial state. For example, for septic patients in ICU, this corresponds to their states upon admission. Although, there has been evidence supporting this way of grouping for different sepsis phenotypes \citep{seymour2019derivation} an interesting avenue of future research is considering groups based on transition dynamics. For example, patients that react differently to various medications may be part of the same group. 
\section*{Acknowledgement}
The research reported here was supported by DEVCOM Army Research Laboratory under Cooperative Agreement W911NF-17-2-0196, and NSF IIS-2007076.

\bibliographystyle{plainnat}
\bibliography{main}

\newpage
\appendix
\section{Appendix}
\subsection{Proof}
Here we present the proof of the theorem \ref{th:loss}.
We restate the theorem first.
\begin{theorem*}[\ref{th:loss}] 
    For a given partition $L\in \Pi$, let $T(x;L)$ be the group treatment
    effect defined in equation \ref{eq:group_treatment_effect}, $t(x)$ be the individual treatment effect as defined in equation \ref{eq:ind_t} and $\hat{T}(x;L)$ an unbiased estimator of
    $T(x;L)$. The following display impose the same ranking over the partitions as the MSE loss in equation \ref{eq:mse}:
    \begin{align}
         -\mathbb{E}_{x\sim\mathcal{X}}\left[\hat{T}^2(x;L)\right] + 2~\mathbb{E}_{x\sim\mathcal{X}}\left[\mathbb{V}\left[\hat{T}(x;L) \right]\right]
    \end{align}
    Where $\mathbb{V}[\hat{T}(x;L)]$ is the variance of the estimator $\hat{T}(x;L)$. 
\end{theorem*}   

\begin{proof}
 
First, We form the adjusted MSE (AMSE) as
\begin{align}
    \nonumber AMSE(\hat{T}; L) = \underset{x\sim\mathcal{X}}{\mathbb{E}} \left[\left(t(x) - \hat{T}(x; L)\right)^2 - t(x)^2\right]
\end{align}
Adjusted MSE and MSE impose the same ranking among different
partitioning as $\underset{x\sim\mathcal{X}}{\mathbb{E}}[t(x)^2]$ is independent from the partitioning. Note that adjusted MSE, similar to MSE cannot be computed.

We continue by decomposing the adjusted MSE by adding
and subtracting $T(x;L)$,
\begin{align*}
    AMSE(\hat{T};L) &= - \mathbb{E}_{x\sim\mathcal{X}}\left[\left(t(x) - \hat{T}(x; L)\right)^2 - t(x)^2 \right]\\
    &=\mathbb{E}_{x\sim\mathcal{X}} [\underbrace{(t(x) - T(x;L))^2- t(x)^2}_{(i)} \\
    &+ \underbrace{(T(x;L)-\hat{T}(x; L))^2}_{ii} \\
    &+\underbrace{2 (t(x) - T(x;L)) (T(x;L) -             \hat{T}(x; L))}_{(iii)}] 
\end{align*}
Now we look at each part separately, for part $(i)$,
\begin{align*}
    &\mathbb{E}_{x\sim\mathcal{X}}\left[(t(x) - T(x;L))^2- t(x)^2\right] = \\
  & \mathbb{E}_{x\sim\mathcal{X}} \left[T(x;L)^2 - 2 t(x) T(x;L)\right]
\end{align*}
Now we expand the expectation over each group of
thr partition $L=\{l_1, \dots, l_M\}$,
\begin{align*}
     \sum_{l_i \in L} P(l_i) T(x;l_i)^2 - 2  \sum_{l_i \in L} P(l_i) T(x;l_i)
\end{align*}
Where $T(x;l_i) = T(x;L)$ such that $x\in l_i$, note
that by definition,  $T(x;l_i)$ is constant for all 
$x\in l_i$. Next, note that $\mathbb{E}_{x\in l_i}[t(x)] = T(x;l_i)$.
\begin{align}\label{app:eq1}
 \nonumber \sum_{l_i \in L} &P(l) T(x;l_i)^2 - 2 \sum_{l_i \in L} P(l_i) T(x;l_i)^2 = \\
      -\sum_{l_i \in L} &P(l_i) T(x;l_i)^2 = -\mathbb{E}_{x\sim\mathbb{X}}[T(x;L)^2]
\end{align}
Now, consider the variance of $\hat{T}(x;l_i)$ for
group $l_i \in L$,

\begin{align}\label{app:eq2}
  \nonumber \mathbb{V} \left[ \hat{T}(x;l) \right] &= \underset{x \in l_i}{\mathbb{E}} [\hat{T}^2(x;l_i) ] - \left[\underset{x \in l_i}{\mathbb{E}}  \hat{T}(x;l_i) \right]^2   \\
  &= \underset{x \in l_i}{\mathbb{E}} [\hat{T}^2(x;l_i) ] - T(x;l_i)^2
\end{align}
Which follows by $\hat{T}(x;l_i)$ being an unbiased
estimator of $T(x;l_i)$. Taking the expectation over
the feature space and substituting equation \ref{app:eq2} into \ref{app:eq1},
\begin{align*}
    (i) &= -\sum_{l_i\in L}P(l_i) T(x;l_i)^2\\
        &= \sum_{l_i\in L}P(l_i) \left[\mathbb{V} \left[ \hat{T}(x;l_i) \right] - \mathbb{E}_{x\sim l_i}\left[ \hat{T}^2(x;l_i) \right] \right]\\
        &=
    \mathbb{E}_{x\sim\mathbb{X}} \left[\mathbb{V} \left[ \hat{T}(x;l) \right]\right] - \mathbb{E}_{x\sim\mathbb{X}}\left[ \hat{T}^2(x;l) \right]
\end{align*}

Now we consider part $(ii)$, 
\begin{align*}
    &\mathbb{E}_{x \sim \mathbb{X}}\left[\left(T(x;L)-\hat{T}(x;L)\right)^2\right] \\
    &= \sum_{l_i}P(l_i) \mathbb{E}_{x\in l_i}\left[ \left(T(x;l_i)-\hat{T}(x;l_i)\right)^2 \right]\\
    &= \sum_{l_i}P(l_i) \mathbb{E}_{x\in l_i}\left[ \left( \mathbb{E}_{x\in l_i} [\hat{T}(x;l_i)]-\hat{T}(x;l_i)\right)^2 \right]\\
    &= \sum_{l_i} P(l_i) \mathbb{V}\left[\hat{T}(x;l_i)\right]\\
    &=\mathbb{E}_{x\sim\mathbb{X}}\left[ \mathbb{V}\left[\hat{T}(x;L) \right] \right]
\end{align*}
Where the third line follows by $\hat{T}(x;l_i)$ being
an unbiased estimator of $T(x;l_i)$.

Looking at the last term $(iii)$,
\begin{align*}
    \mathbb{E}_{x\sim\mathcal{X}}\left[\left(t(x)\hat{T}(x;L)\right)\right] &= \sum_{l_i\in L}P(l_i) \mathbb{E}_{x\in l_i}\left[t(x)\hat{T}(x;l_i)\right]\\
    &= \sum_{l_i\in L}P(l_i) \hat{T}(x;l_i) \mathbb{E}_{x\in l_i}\left[t(x)\right]\\
    &= \sum_{l_i\in L}P(l_i) \hat{T}(x;l_i) T(x;l_i)\\
    &= \mathbb{E}_{x\in\mathcal{X}}\left[\hat{T}(x;l_i) T(x;l_i)\right]
\end{align*}
Which implies $\mathbb{E}_{x\sim\mathcal{X}} \left[(t(x) - T(x;L))\hat{T}(x;L) \right] = 0$. As a result,
\begin{align*}
    (iii) &= 2\mathbb{E}_{x\in \mathcal{X}}\left[ \left(t(x)-T(x;L)\right)\left(T(x;L)-\hat{T}(x;L) \right) \right]\\
    &= 2\mathbb{E}_{x\in \mathcal{X}}\left[ \left(t(x)-T(x;L)\right)\left(T(x;L)\right) \right]\\
    &-2\mathbb{E}_{x\in \mathcal{X}}\left[ \left(t(x)-T(x;L)\right)\left(\hat{T}(x;L)\right) \right] = 0
\end{align*}
Putting the results together, results in equation \ref{eq:loss}.
\end{proof}

Next, we continue with the proof of Theorem \ref{th:variance}. First, we present a reminder on Rényi divergence. 
\paragraph{Rényi Divergence}  For $\alpha \geq 0$ the Rényi divergence for two distribution $P$ and $Q$ as defined by \citep{cortes2010learning} is 
\begin{align*} 
D_\alpha(P||Q) =\frac{1}{\alpha-1}\log_2\sum_x Q(x)\left(\frac{P(x)}{Q(x)}\right)^{\alpha - 1}.
\end{align*}

Denote the exponential in base 2 by $d_\alpha(P_e||P_b) = 2^{D_\alpha(P_e||P_b)}$.

The effective sample size (ESS) \citep{kong1992note} is often
used for diagnosis of IS estimators, and is defined as 
\begin{align*}
    ESS(P||Q) =\frac{N}{1+\mathbb{V}_{x\sim Q}[w(x)]} = \frac{N}{d_2(P||Q)}
\end{align*}

where $N$ is the number of samples drawn to estimate the importance weights. A common estimator of the ESS\cite{mcbook} is 
\begin{align*}
\widehat{ESS}(P||Q) = \frac{\left(\sum_{i=1}^N w_i\right)^2}{\sum_{i=1}^Nw_i^2}
\end{align*}

\begin{theorem*}[\ref{th:variance}]
Given a dataset $\mathcal{D}=\{(x_0, \rho_0, g_0),\dots,(x_N,\rho_N,g_N)\}$ and the treatment effect estimator defined by $\hat{T} = \frac{1}{N}\sum_i(\rho_i - 1)g_i$. The variance of $\hat{T}$ satisfies the following inequality,
\begin{align}\label{eq:upp_bound_appx}
    \mathbb{V}[\hat{T}] \leq \left\lVert g\right\lVert^2_\infty\left(\frac{1}{ESS} - \frac{1}{N}\right)
\end{align}
\noindent where, $ESS$ is the effective sample size.
\end{theorem*}

\begin{proof}
First note that the variance of the treatment effect 
estimator $\hat{T} = \frac{1}{N}\sum_i(\rho_i - 1)g_i$ can be
upper bounded by the variance of the importance sampling weights. Since
$\mathbb{V}[\hat{T}] \leq \mathbb{E}[\hat{T}^2]$

\begin{align*}
    \mathbb{V}[\hat{T}] \leq \frac{||g||^2_\infty}{N^2} \mathbb{E}\left[\sum_{i}(\rho_i - 1)^2\right] = \frac{1}{N}||g||^2_\infty \mathbb{V}[\rho],
\end{align*}
where the last equality follows by observing that $\mathbb{E}[\rho]=1$. As noted by~\citet{metelli2018policy},  The variance of the treatment effect estimator can be written as 
\begin{align*}
    \mathbb{V}[\hat{T}] \leq \||g||^2_\infty (\frac{d_2(P_e||P_b)}{N} - \frac{1}{N})
\end{align*}
This expression can be related to the effective sample size of the original dataset given the evaluation policy, resulting in equation \ref{eq:upp_bound_appx}
\begin{equation*}
    \mathbb{V}[\hat{T}] \leq ||g||^2_\infty\left(\frac{1}{ESS} - \frac{1}{N}\right)
\end{equation*}

\end{proof}
\section{Experimental Details}
In this section we present details of the experimental setups along
with some extra experiments that shows the effect
of the regularization on the results reported in the
main text.
\subsection{Simple Illustration}
We consider a simple Markov
decision process (MDP) with the state space $x \in [0, 1]$, discrete action space
$a \in \{-1, 0, 1\}$ and the reward function is defined as $r(x) = 1-|x-0.5|$.
The transition dynamic is specified by, $x_{t+1} = clip(x_t + \kappa \times a_t + \epsilon, 0, 1)$, where the function $clip(x, a, b)$, clips the value of
$x$ between $a$ and $b$, $\kappa = 0.2$ and $\epsilon \sim \mathcal{N}(0,0.05)$. Each episode lasts $H$ steps.
Intuitively, action $1$ takes the agent to the right, $-1$ to the left and $0$ same location with some gaussian noise.
If the agent hits the boundary, the action has no effect
on the position. 

The behaviour policy, takes action with the following probabilities
\begin{equation*}
\left\{
\begin{aligned}
    x < 0.2:~& \pi_b(-1) = 0.25,~ \pi_b(0) =0.25,~ \pi_b(1)=0.5\\
    x \geq 0.2:~& \pi_b(-1) = 0.5,~ \pi_b(0) =0.25,~ \pi_b(1)=0.25\\
\end{aligned}
\right.
\end{equation*}
And the evaluation policy,
\begin{equation*}
\left\{
\begin{aligned}
    x > 0.8:~& \pi_e(-1) = 0.5,~ \pi_e(0) =0.25,~ \pi_e(1)=0.25\\
    x \leq 0.8:~& \pi_e(-1) = 0.25,~ \pi_e(0) =0.25,~ \pi_e(1)=0.5\\
\end{aligned}
\right.
\end{equation*}

\begin{figure}[tb]
    \begin{tabular}{cc}
    \includegraphics[width=0.45\linewidth]{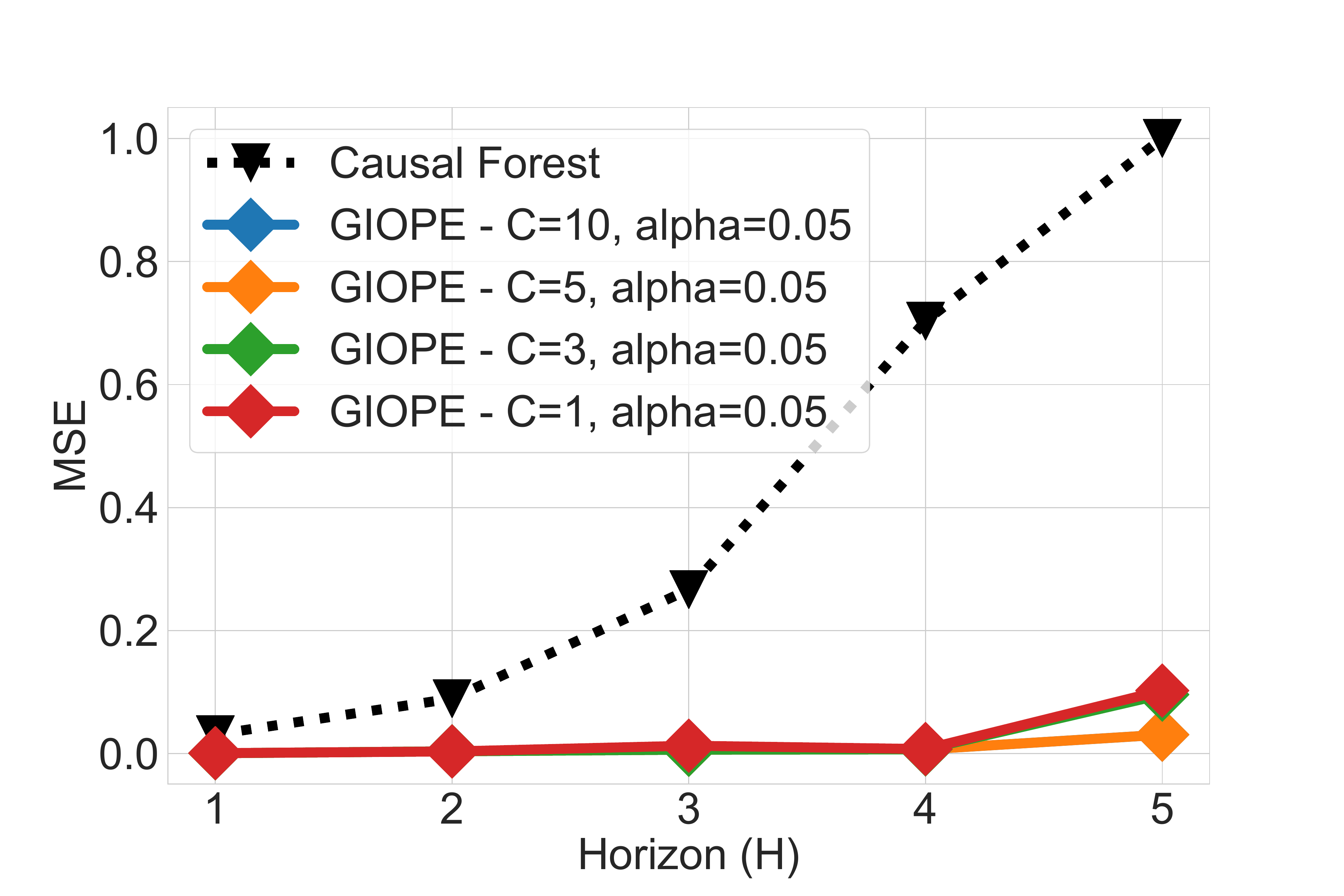} &
    \includegraphics[width=0.45\linewidth]{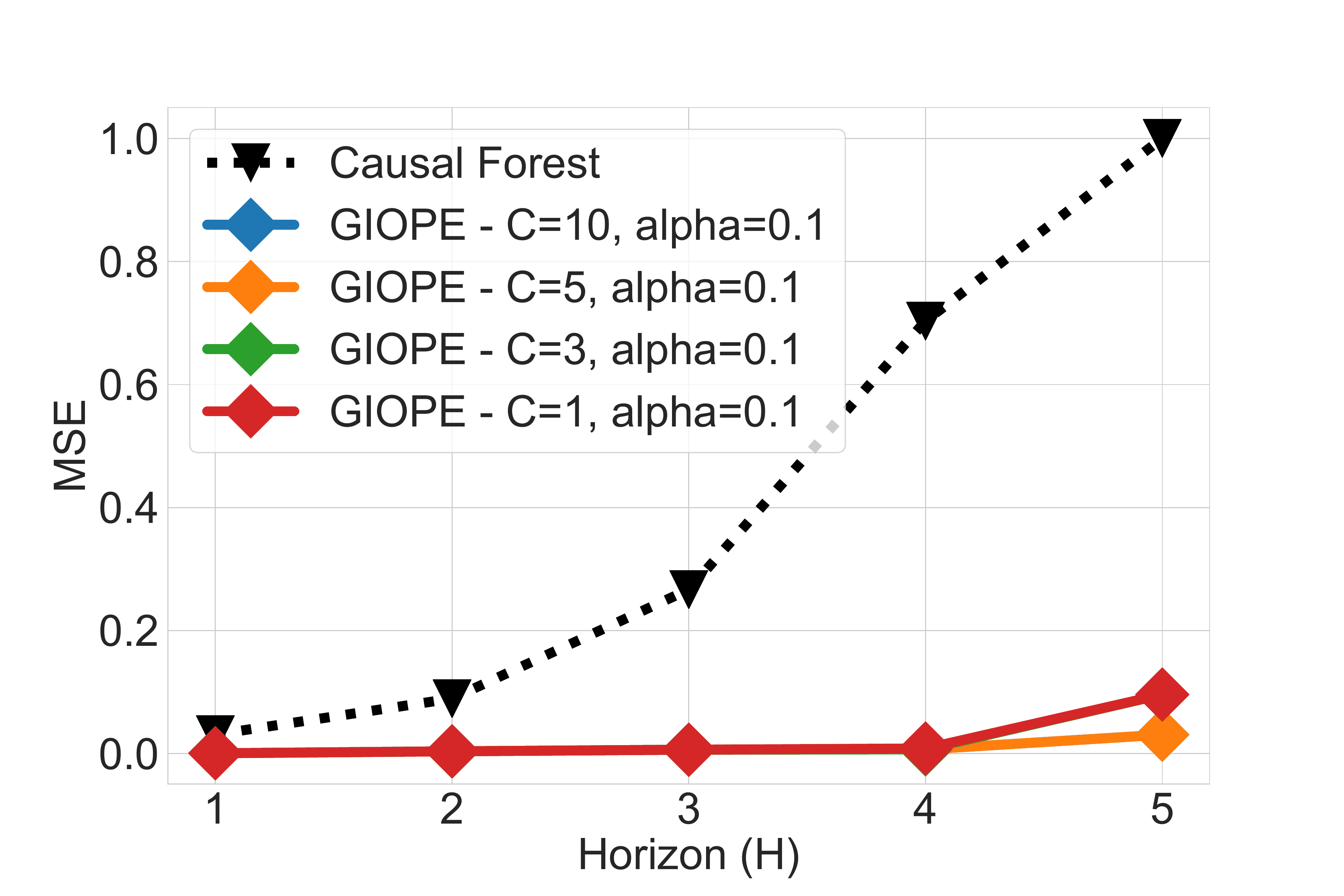}\\
    (a) $\alpha=0.05$ &  (b) $\alpha=0.1$
\end{tabular}
\caption{Toy MDP. (a) regularization margin $\alpha = 0.05$, (b) regularization margin $\alpha = 0.1$}\label{fig:toy_reg}
\end{figure}

We generated $50000$ trajectories
with the behaviour policy for horizons $\{1, 2, 3, 4, 5\}$
and averaged all results over 10 runs. Figure \ref{fig:toy_reg} compares the mean squared
error of our method versus the causal forests
for different range of hyper-parameters. Panel (a)
shows the results for margin $\alpha=0.05$ and
values of regularization constant $C=\{1, 3, 5, 10\}$ and panel (b) shows the results for margin $\alpha=0.1$. As shown, regularization has small effects on the results and the results reported in the main text
holds for a large range of hyper-parameters.

\subsection{Sepsis Simulation}
\begin{figure*}[tb]
    \centering
    \begin{tabular}{cccc}
    \includegraphics[width=0.23\linewidth]{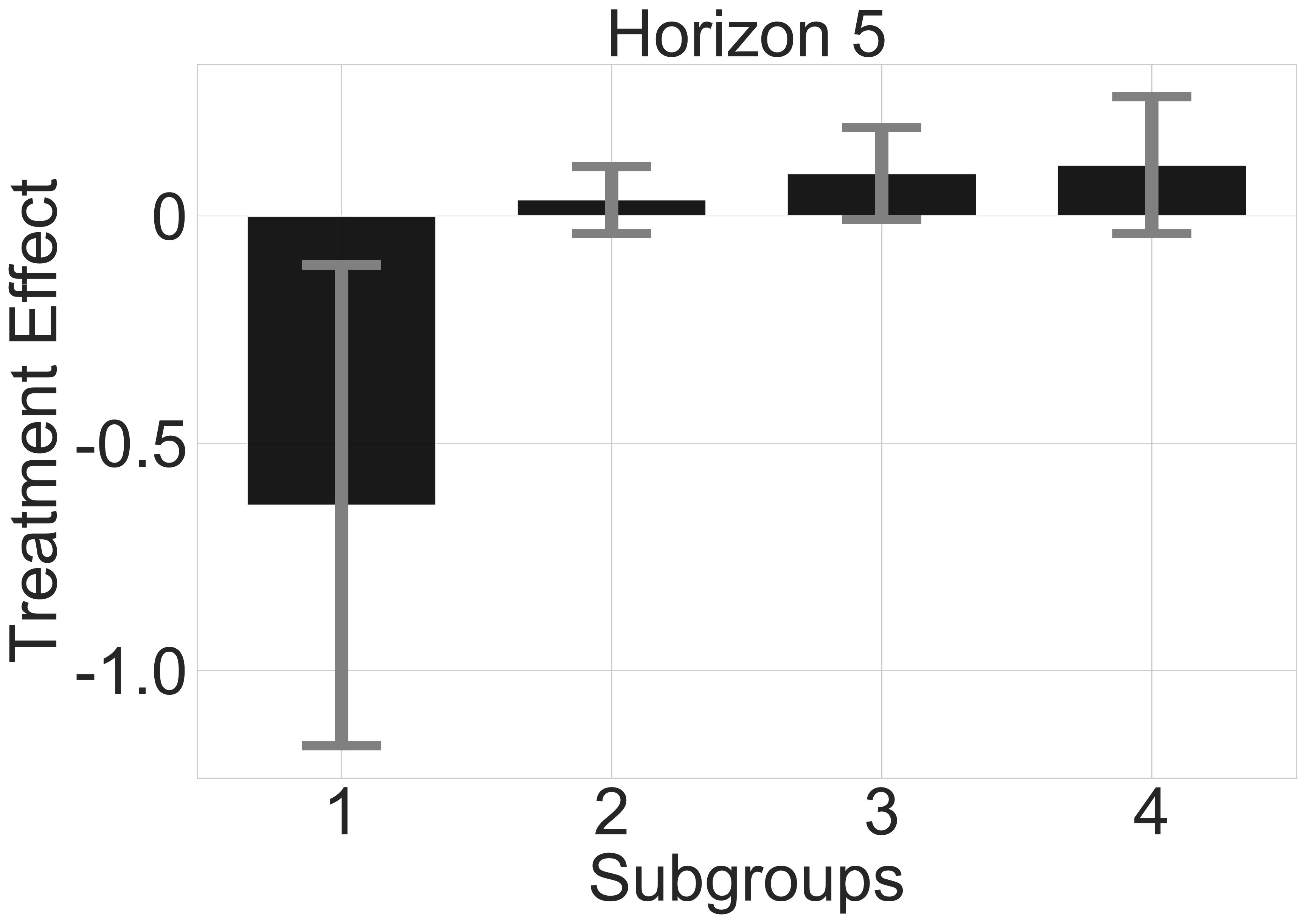} &
    \includegraphics[width=0.23\linewidth]{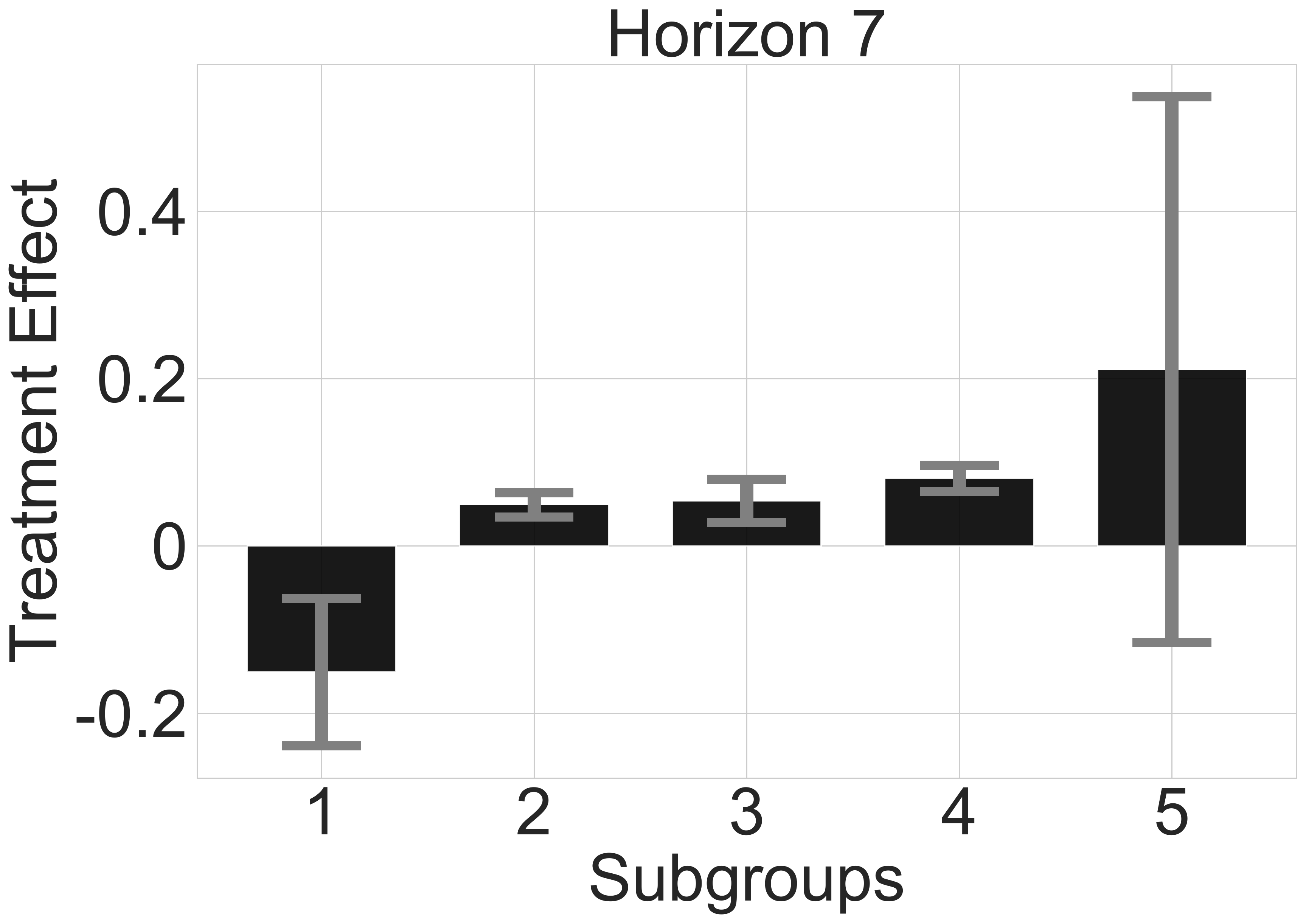} &
    \includegraphics[width=0.23\linewidth]{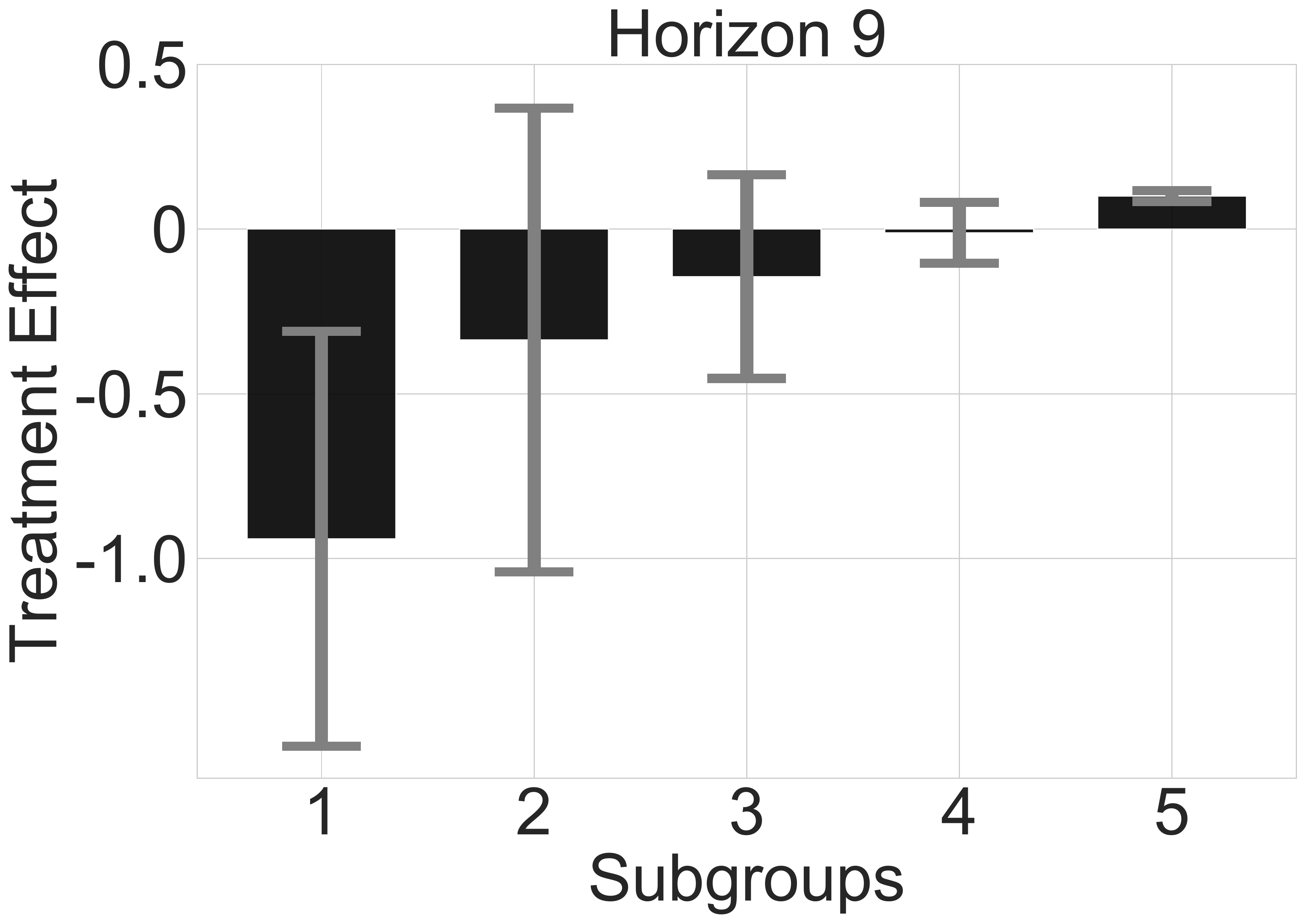} &
    \includegraphics[width=0.23\linewidth]{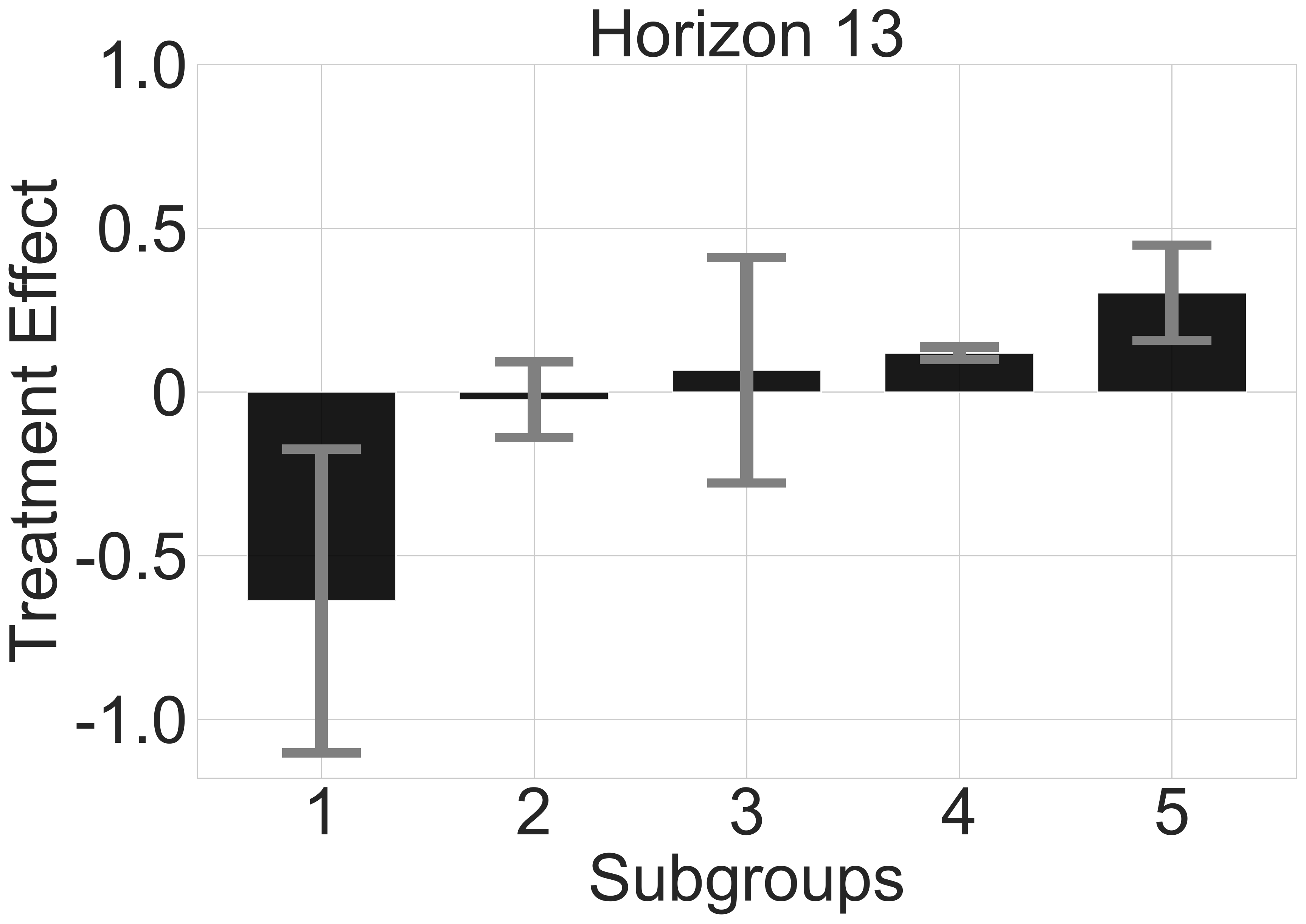} \\
    (a) &  (b) & (c) & (d)\\
\end{tabular}
\caption{Sepsis simulation. Group treatment effect for a sample run with horizon (a) 5, (b) 7, (c) 9 (d) 13}\label{fig:sepsis_all_group}
\end{figure*}

We used the following set of hyper-parameters for the
experiments presented in section \ref{sec:experiments}.
Regularization constant $C=5.0$,
regularization margin $\alpha=0.05$, regularization confidence value
$\delta=0.4$, maximum depth of the tree $d=\infty$  and minimum number of samples in each leaf $50$. Figure \ref{fig:sepsis_all_group} shows the partitions found in horizon $\{5, 9, 11\}$.
As shown, our method
can recover groups with significant negative treatment
effect in every horizon.

\begin{figure}[tb]
    \begin{tabular}{cc}
    \includegraphics[width=0.45\linewidth]{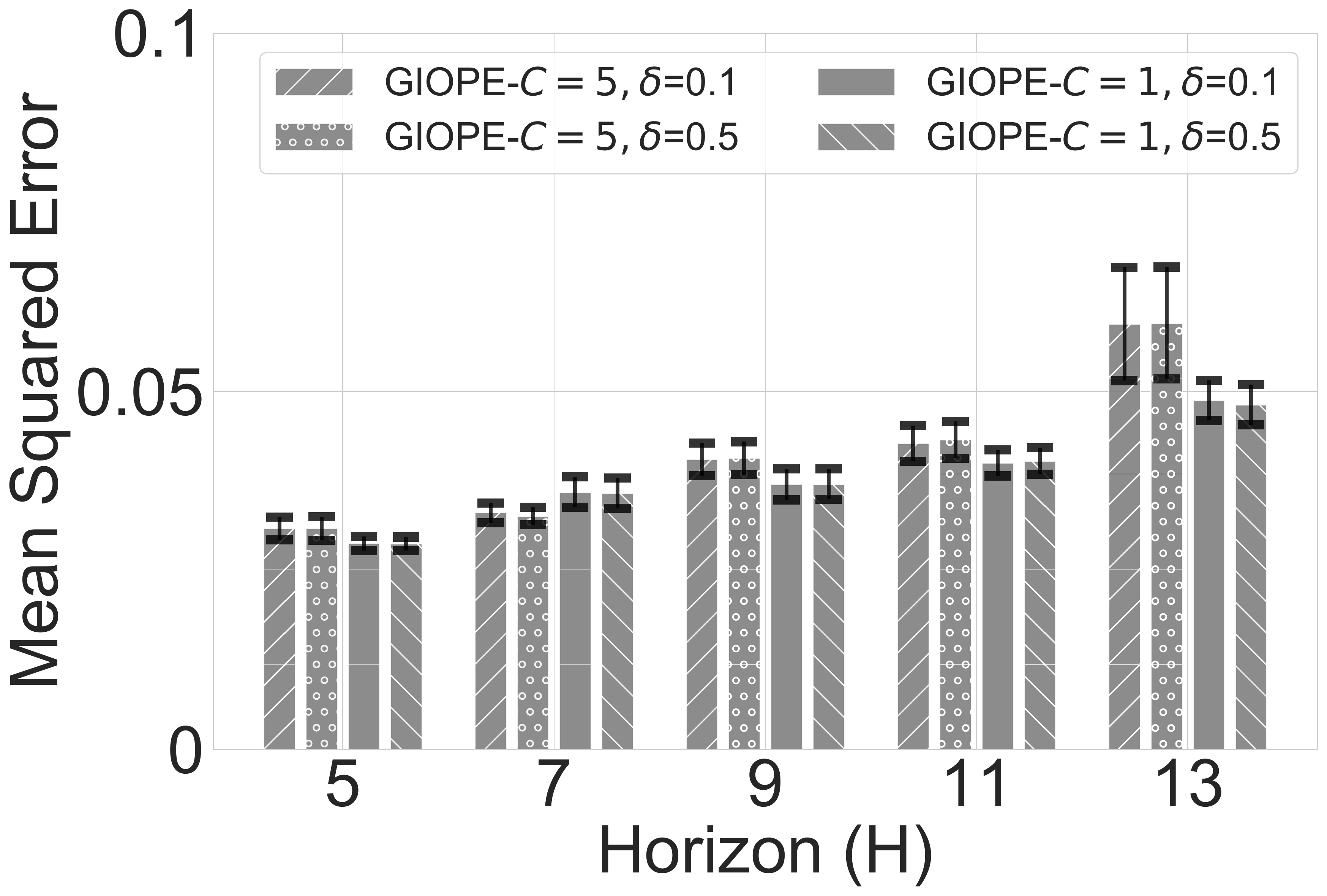} &
    \includegraphics[width=0.45\linewidth]{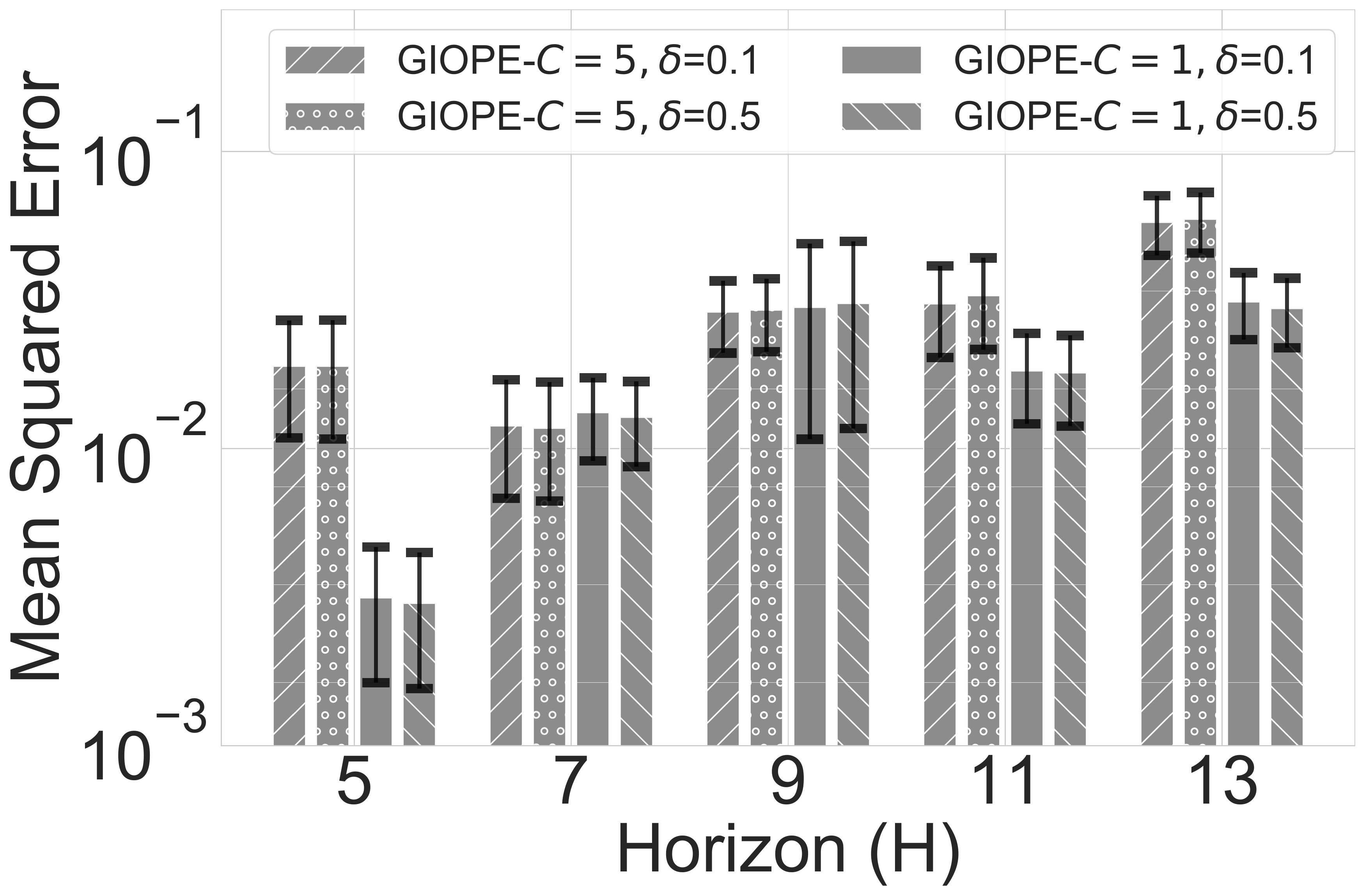}\\
    (a) &  (b) \\
    \includegraphics[width=0.45\linewidth]{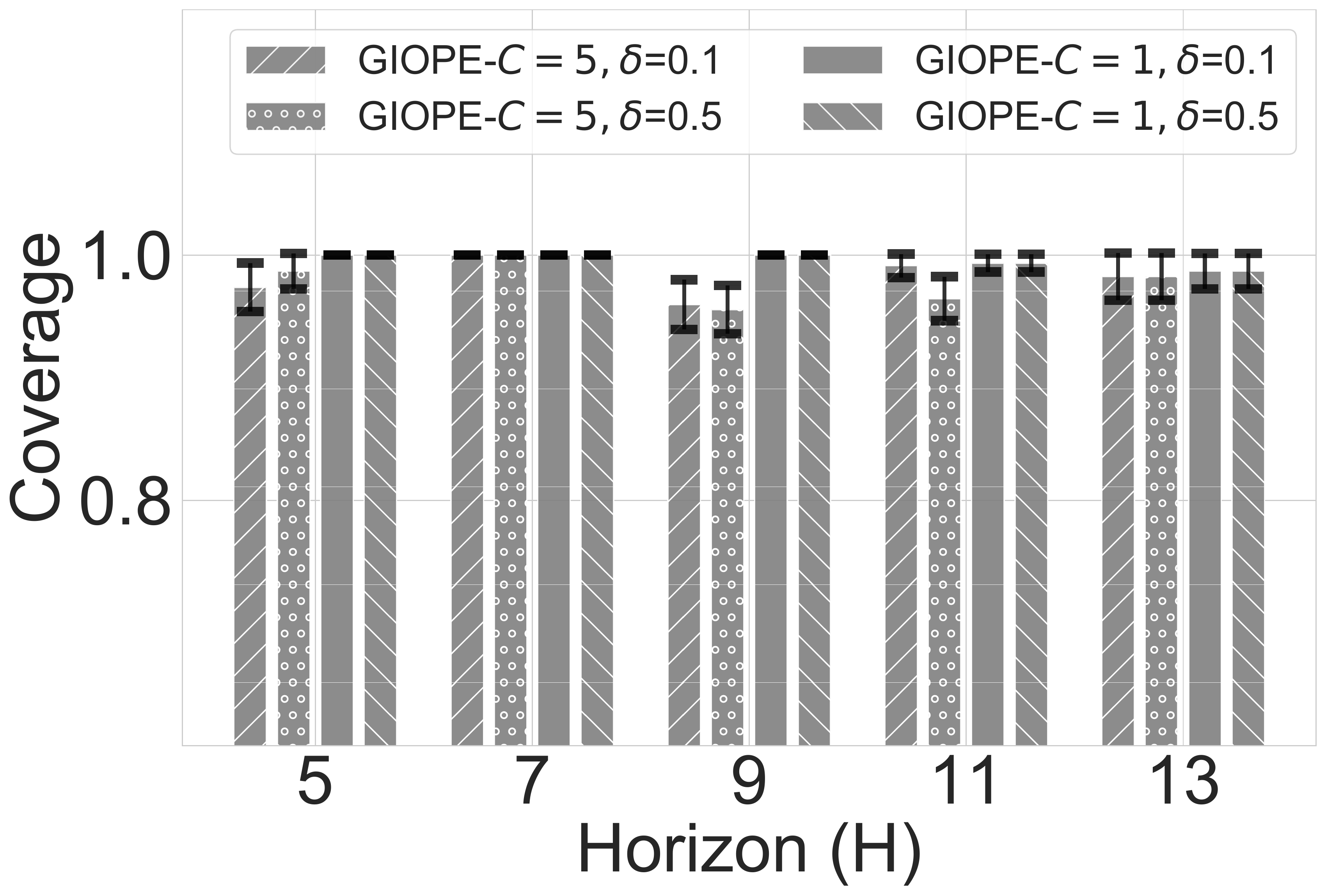} &
    \includegraphics[width=0.45\linewidth]{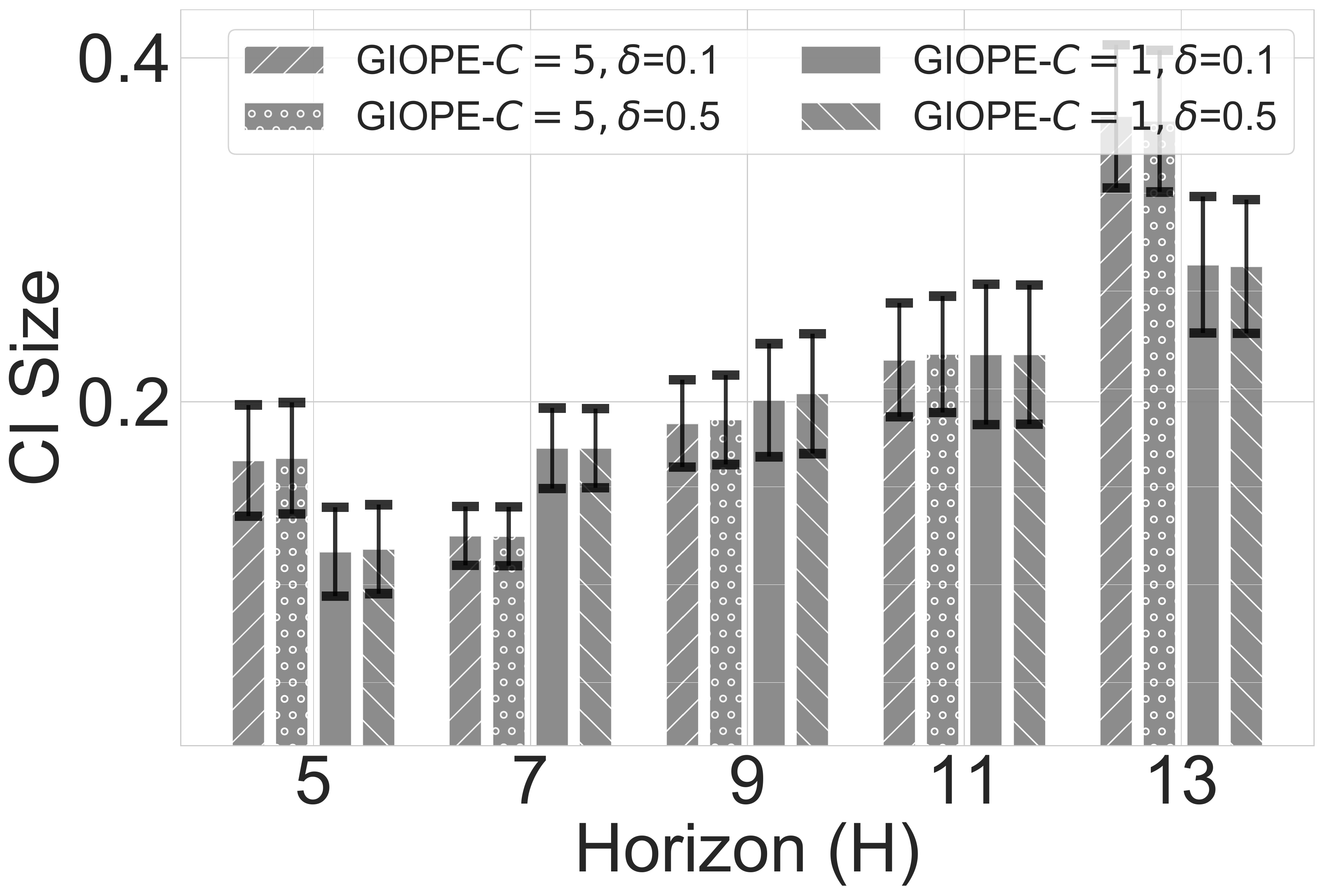}\\
    (c) &  (d) \\
\end{tabular}
\caption{Ablation study, results of GIOPE for four different
values of parameters. (a) Mean squared error (b) group mean squared error, (c) 95\% confidence interval coverage and (d) average size of confidence intervals}\label{fig:ablation_compare}
\end{figure}

In order to evaluate the effect of hyper-parameters, we
perform the ablation study for two different values
of regularization confidence interval $\delta=0.1$ and
$\delta=0.4$ and two different values of regularization constant $C=5.0$ and $C=1.0$. Figure \ref{fig:ablation_compare} (a) shows the result for mean squared error and (b) for group mean squared error. As shown,
the effect of regularization is small, and the same 
results as in the main text can be obtained with 
different range of hyper-parameters. Similarly, figure \ref{fig:ablation_compare} (c) shows the coverage of the 95\% confidence interval and (d) is the average size of CI. The results obtained
in the main text holds with different value
of hyper-parameters.

\subsection{MIMIC-III}
We used MIMIC-III dataset \citep{johnson2016mimic}, and following
\citet{komorowski2018artificial} extracted the sepsis cohort. 
Our training set consist of 14971 individuals, with 8442 male and 6529
female. The mortality rate in our cohort is 18.4\%. The feature
space is of size 44 consist of the following values:

\{\texttt{gender}, \texttt{re\_admission}, \texttt{mechvent},
\texttt{age}, \texttt{Weight\_kg}, \texttt{GCS}, \texttt{HR},
\texttt{SysBP}, \texttt{MeanBP}, \texttt{DiaBP}, \texttt{RR},
\texttt{Temp\_C}, \texttt{FiO2\_1}, \texttt{Potassium}, \texttt{Sodium},
\texttt{Chloride}, \texttt{Glucose}, \texttt{Magnesium},
\texttt{Calcium}, \texttt{Hb}, \texttt{WBC\_count},
\texttt{Platelets\_count}, \texttt{PTT}, \texttt{PT},
\texttt{Arterial\_pH}, \texttt{paO2}, \texttt{paCO2},
\texttt{Arterial\_BE}, \texttt{Arterial\_lactate}, \texttt{HCO3},
\texttt{Shock\_Index}, \texttt{Shock\_Index}, \texttt{PaO2\_FiO2},
\texttt{cumulated\_balance}, \texttt{SOFA}, \texttt{SIRS},
\texttt{SpO2}, \texttt{BUN}, \texttt{Creatinine}, \texttt{SGOT},
\texttt{SGPT}, \texttt{Total\_bili}, \texttt{INR}, \texttt{output\_total},
\texttt{output\_4hourly}\}

We provide the index of the patients
in the dataset to facilitate reproducibility of our results.

In order to estimate the behaviour policy, we use KNN with $k=100$
on the test set, we use $l_2$ distance
with uniform weights across different features to measure the distance. If an action
was not taken among all 100 nearest neighbours, we assign the 
probability $0.01$ to the action. We used
IV fluid and mechanical ventilation for actions and used $20\%$
quantile to discretize the action space into $25$ actions.

For the evaluation policy, we used a similar method as the behaviour 
policy on a random subset of training set (20\% of the training data).
We only used the following features to estimate the distance
for the evaluation policy,

\{\texttt{HR}, \texttt{SysBP}, \texttt{Temp\_C}, \texttt{Sodium},
\texttt{Chloride}, \texttt{Glucose}, \texttt{Calcium},
\texttt{paO2}, \texttt{Arterial\_BE}, \texttt{SOFA},
\texttt{SIRS}, \texttt{Creatinine}\}

Simiarly, if an action
was not taken among all 100 nearest neighbours, we assign the 
probability $0.01$ to the action. In our experiments in section \ref{sec:experiments} we used the 
following set of hyper-parameters: regularization constant $C=100.0$,
regularization margin $\alpha=0.0$, regularization confidence value
$c=2.0$, maximum depth of the tree $d=\infty$ and minimum number of samples in each leaf $1000$.
       

\end{document}